\documentclass[letterpaper]{article} 
\usepackage{aaai24_arxiv}  
\usepackage{times}  
\usepackage{helvet}  
\usepackage{courier}  
\usepackage[hyphens]{url}  
\usepackage{graphicx} 
\usepackage{tabularx}
\usepackage{multirow}
\usepackage{booktabs}
\usepackage{pifont}
\newcommand{\xmark}{\ding{55}}%
\usepackage{natbib}  
\usepackage{caption} 
\usepackage{algorithm}
\usepackage{algorithmic}
\usepackage{amsmath,amssymb,amsfonts}
\usepackage{xspace}
\usepackage{tikz}
\usepackage{subcaption}
\usetikzlibrary{shapes.geometric}
\DeclareRobustCommand\diamonds[1]{\tikz[baseline=(char.base)]{
            \node[shape=diamond,draw,inner sep=0pt] (char) {\textcolor{black}{#1}};}}
\DeclareMathOperator{\sstop}{stop}
\DeclareMathOperator{\moveforward}{move\_forward}
\DeclareMathOperator{\turnleft}{turn\_left}
\DeclareMathOperator{\turnright}{turn\_right}

\newcommand{\stateset}{\mathcal{S}}
\newcommand{\actionset}{\mathcal{A}}
\newcommand{\Value}{V}

\newcommand{\expect}{\mathbb{E}}
\newcommand{\real}{\mathbb{R}}
\newcommand{\policy}{\pi}
\newcommand{\discount}{\gamma}
\newcommand{\Oprobs}{\mathcal{Z}}
\newcommand{\Reward}{R}
\newcommand{\reward}{r}
\newcommand{\observations}{\Omega}
\newcommand{\transition}{T}

\newcommand{\Belief}{b}

\newcommand{\state}{s}
\newcommand{\action}{a}
\newcommand{\obs}{o}

\newcommand{\goal}{g}
\newcommand{\Goal}{G}

\newcommand{\pigoal}{\pi_g}
\newcommand{\pivln}{\pi_\ell}
\newcommand{\piquestion}{\pi_{ques}}
\newcommand{\pis}{\pi_s}

\newcommand{\memory}{M}

\newcommand{\zetavln}{\zeta_\ell}
\newcommand{\zetaquestion}{\zeta_{ques}}

\newcommand{\Question}{\mathcal{Q}}
\newcommand{\Instruction}{\mathcal{I}}

\newcommand{\Evaluator}{\mathcal{E}}
\newcommand{\QuesGenerator}{\mathcal{G}^q}
\newcommand{\InsGenerator}{\mathcal{G}^i}

\newcommand{\set}[1]{\left\{#1\right\}}

\newcommand{\card}[1]{\left|{#1}\right|}

\newcommand{\name}{CAVEN\xspace}
\newcommand{\etal}{et al.\xspace}

\usepackage{newfloat}
\usepackage{listings}
\DeclareCaptionStyle{ruled}{labelfont=normalfont,labelsep=colon,strut=off} 
\floatstyle{ruled}
\newfloat{listing}{tb}{lst}{}
\floatname{listing}{Listing}

\begin{document}

\title{CAVEN: An Embodied Conversational Agent for \\Efficient Audio-Visual  Navigation in Noisy Environments}

\author {
    Xiulong Liu\textsuperscript{\rm 1}\thanks{Work done while interning at MERL.},
    Sudipta Paul\textsuperscript{\rm 2}\thanks{Work done while the author was at UC Riverside.},
    Moitreya Chatterjee\textsuperscript{\rm 3},
    Anoop Cherian\textsuperscript{\rm 3}
}
\affiliations {
    \textsuperscript{\rm 1}University of Washington, Seattle, WA\quad
    \textsuperscript{\rm 2}Samsung Research America, Mountain View, CA\\
    \textsuperscript{\rm 3}Mitsubishi Electric Research Labs, Cambridge, MA\\
    xl1995@uw.edu, spaul007@ucr.edu, metro.smiles@gmail.com, cherian@merl.com
}

\maketitle

\begin{abstract}
Audio-visual navigation of an agent towards locating an audio goal is a challenging task especially when the audio is sporadic or the environment noisy. In this paper, we present \name,  a Conversation-based Audio-Visual Embodied  Navigation framework in which the agent may interact with a human/oracle for solving the task of navigating to an audio goal. Specifically, \name is modeled as a budget-aware partially observable semi-Markov decision process that implicitly learns the uncertainty in the audio-based navigation policy to decide when and how the agent may interact with the oracle. Our \name agent can engage in fully-bidirectional natural language conversations by producing relevant questions and interpret free-form, potentially noisy responses from the oracle based on the audio-visual context. To enable such a capability, \name is equipped with: (i) a trajectory forecasting network that is grounded in audio-visual cues to produce a potential trajectory to the estimated goal, and (ii) a natural language based question generation and reasoning network to pose an interactive question to the oracle or interpret the oracle's response to produce navigation instructions.  To train the interactive modules, we present a large scale dataset: AVN-Instruct, based on the Landmark-RxR dataset. To substantiate the usefulness of conversations, we present experiments on the benchmark audio-goal task using the SoundSpaces simulator under various noisy settings. Our results reveal that our fully-conversational approach leads to nearly an order-of-magnitude improvement in success rate, especially in localizing new sound sources and against methods that only use uni-directional interaction.
\end{abstract}

\section{Introduction}
\begin{figure}[t]
    \centering
  \includegraphics[width = \linewidth]{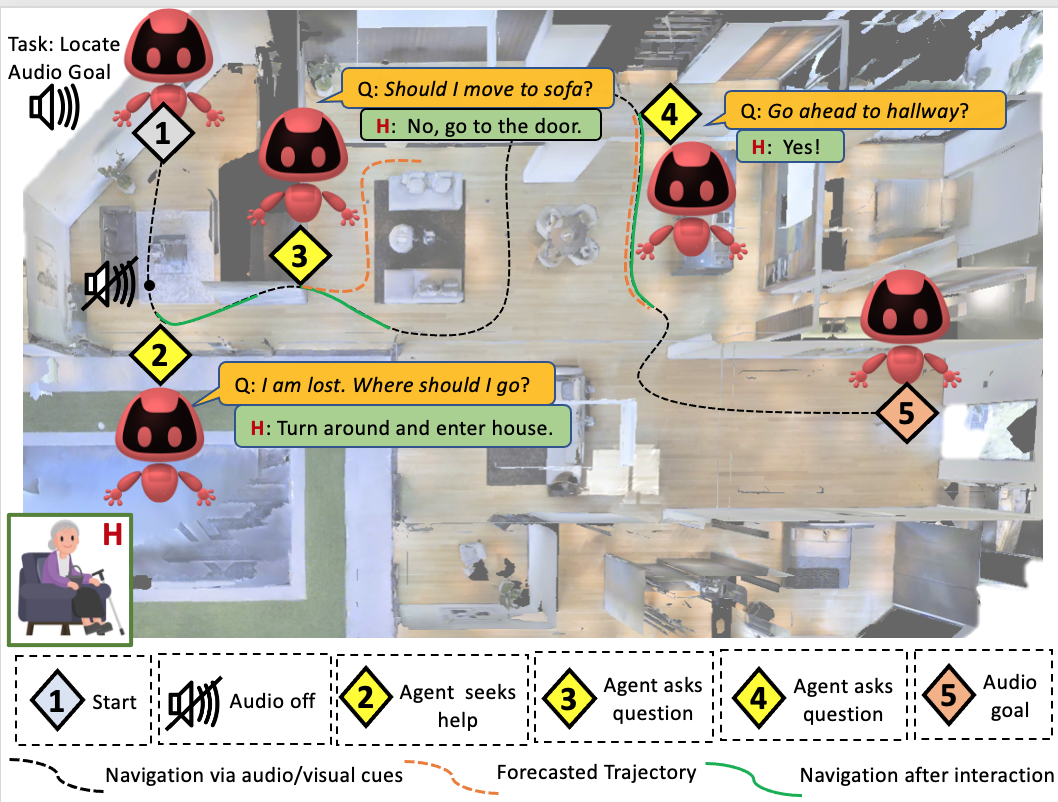}
  \caption{An illustrative \name interaction: The agent starts at \diamonds{1} guided by the audio event at \diamonds{5}. At \diamonds{2}, the agent decides to seek help from the human/oracle H (e.g., because the audio stopped). The oracle then provides a short natural language instruction for the agent to follow. At locations \diamonds{3} and \diamonds{4}, the agent decides to ask questions to the oracle using the forecasted trajectories (orange) and gets feedback, finally reaching the audio goal at \diamonds{5}. }
  \label{fig:task}
\vspace{-6mm}
\end{figure}

The advent of powerful deep neural networks and sophisticated language models have led to significant advancements in building conversational agents that can collaborate with humans in solving challenging reasoning tasks~\cite{peng2023instruction,ram2018conversational,chowdhery2022palm,gupta2023visual,you2022end}. While, much effort has been expended on tasks that are predominantly in the language domain, such progress is yet to percolate into real world problems that need complex reasoning over multiple modalities of perception~\cite{li2022learning,liu2023tackling}. One such task that we exclusively explore in this paper is that of audio-visual navigation of an embodied robotic agent where the goal is to localize a sound producing source in a realistic, complex, and never-seen before environment when the sound is noisy, intermittent, sporadic, and mixed with other sounds — a situation even humans may find hard to tackle. As can be easily imagined, the applications needing such an \emph{audio goal} capability are enormous; for example, at one end, we may think of a robotic disaster and emergency response agent that may need to move through a huge rubble to localize victims of an earthquake who may cry for help and on the other, one may consider a home robotic vacuum repurposed to be vigilant to strange sounds. 

While the task of navigating to the audio goal, has witnessed some attention in the recent years (Chen \etal, 2021), we consider a variant of this task, dubbed audio-visual-language embodied navigation (AVLEN) (Paul \etal, 2022), where the agent has the ability to interact with a human/oracle when it is unable to solve the task by itself and potentially query an oracle for task-specific guidance. However, the interactive abilities of the agent in AVLEN is limited in several aspects. In particular, the AVLEN agent could ask only a fixed question (e.g., ``Help me!"), while the (human) oracle could provide a natural language response for guiding the agent to the goal. This technique of querying, while useful to some extent, does not cover the full scope of bi-directional interactions. As we know, back and forth interaction in natural language simulates a human-like conversation, allowing for better expressivity towards sharing ideas effectively. For instance, let's assume for a moment that the agent is a 5 year old child who needs help in finding a sounding toy at a secret location. While the parents (oracle) could suggest: ``look inside the wooden trunk'' (as in Paul \etal, 2022), the child might not know what a `trunk' is. Instead, isn't it better if the child had asked: "Should I look next to the large brown box?" and the parents say: "yes"? or suggest "No, look inside it"? It is not only easy to respond with a `yes'/`no' answer (if possible), but this also avoids the need to know what a `trunk' is (and ask more questions or make wrong inferences). Engaging in conversations to resolve such ambiguities is of importance in several time-critical real-world circumstances, e.g., the sound of wheezing in an elderly care or a thud in a medical facility.

Our goal in this paper is to build a fully-conversational robotic agent, which we call \name~-- Conversational Audio-Visual Embodied Navigation,  with the capabilities as described above, that can engage in bidirectional interactions with an oracle in natural language towards solving the audio goal task in a complex realistic visual environment. Specifically, \name can either use the audio-visual cues for its navigation (as in prior works~\cite{chen2020soundspaces, chen2021learning,gan2020look}) or in case the agent is uncertain of which navigation step to take, it can interact with the oracle in two distinct modes: 
(i) a \emph{question mode}, in which the agent forecasts a plausible trajectory based on audio-goal belief, using which it frames a natural language question to be posed to the oracle, and subsequently interpreting the oracle's response to the question, and (ii) a \emph{query mode}, where the agent is unsure of what question to even phrase (e.g., when there are no useful cues in the scene) or completely uncertain about its current situation, and therefore directly seeks the oracle's guidance. Figure~\ref{fig:task} illustrates a typical conversation between a human and our agent.

There are several challenges to tackle when designing the learning and inference model for \name. Specifically, (i) when should the agent use language? (ii) what type of language interaction should the agent use (question or query)? (iii) how should the agent phrase the question? (iv) how to make the oracle understand the agent’s question?, (v) how should the oracle respond to the agent’s question? and (vi) how frequently  should the agent be allowed to ask questions (budget)? Note that, some of these challenges are partially addressed in prior works~\cite{kesiraju2020learning,siddhant2018deep,xiao2019quantifying} such as (v) and (vi). However in \name, we tackle all these challenges within a single framework, by proposing a novel budget-aware partially observable semi-Markov decision process (POSMDP), using a reinforcement learning framework by introducing novel learning rewards. 

To empirically assess the performance of \name, we conduct extensive experiments on the semantic audio-goal navigation task~\cite{chen2021semantic} in the SoundSpaces simulator, under various challenging scenarios, each having intermittent sounds emanating from a source. One key difficulty to train the \name model is the absence of any large scale dataset that includes language instructions in an audio-visual navigation setting. To this end, we introduce \emph{AVN-Instruct} -- a novel audio-visual-language navigation sub-instruction dataset with 41.5k pairs of audio-goal, trajectory, and language instructions. 
Our experimental results using the above setup clearly bring out the benefits of enabling the agent to converse with the oracle, demonstrating a solid gain of nearly $12\%$ over competing approaches on the success rate. 

We summarize below the core contributions of our work:
\begin{itemize}
    \item  We present \name, a multimodal navigation agent that is, for the first time,  capable of fully-bidirectional interaction with an oracle in free-form natural language, thereby facilitating easy communication. 
    \item We introduce a novel \emph{question module} for bi-directional interaction with the oracle consisting of: (i) a trajectory forecasting module grounded on both visual scenes and audio cues, (ii) a question generation module, and (iii) a question decoder (FollowerNet, on the oracle side).
    \item We design a novel budget-aware and uncertainty-splitting reinforcement learning policy, which integrates the question module as an additional policy (using suitable reward design inspired by differential RL) in addition to audio-visual navigation and language-based policy. 
    \item We propose a novel audio-visual-language navigation sub-instruction dataset, {AVN-Instruct} to pre-train embodied navigation models. We also propose two new metrics to evaluate language-guided navigation tasks, dubbed SNO and SNI. 
    \item Our experiments demonstrate state-of-the-art performances against related prior approaches by an order-of-magnitude increase in success rate. 
\end{itemize}

\section{Related Works}
\textbf{Audio-Visual Embodied Navigation Tasks:} Recent years have seen several works in Embodied AI that consider the audio-goal navigation task~\cite{chen2020soundspaces, chen2021learning,gan2020look,yu2022sound}. Generally, this task assumes a continuous sound. However, there are derivatives that look at situations when the audio is sporadic and depends on the category of the sounding object, dubbed semantic audio-goal navigation~\cite{chen2021semantic}. Both of these tasks are facilitated by the SoundSpaces simulator~\cite{chen2020soundspaces} that can render realistic audio in 3D visual environments. While the aforementioned methods only consider audio and visual modalities, (Paul \etal, 2022) proposes AVLEN that utilizes language feedback from the oracle. However, there is no provision of posing questions, which burdens the oracle with the task of chalking out a path to the goal whenever help is sought. Contrary to these approaches, our proposed \name utilizes bi-directional interaction with the oracle besides audio-visual cues, a setting that is more practical. 

\noindent \textbf{Vision-and-Language Navigation (VLN):} The task in VLN is to use (or execute)  natural language instructions to reach a target location. Akin to~\cite{gu2022vision}, we group VLN approaches in three categories: (i) instruction-at-start, (ii) oracle guidance, and (iii) bi-directional interaction. \emph{Instruction-at-start} is a well-explored research area~\cite{anderson2018vision, Hong_2021_CVPR, ke2019tactical, liu2021vision, majumdar2020improving, ma2019self, ma2019regretful, zhu2020babywalk, chen2021history, pashevich2021episodic, guhur2021airbert} in which the agent is given a language instruction at its start describing the intended path. To tackle the task, Wang \etal~\cite{wang2019reinforced} uses cross-modal attention to focus on the relevant parts of both vision and language modalities, while others~\cite{fried2018speaker,tan2019learning}, used augmented instruction-trajectory pairs to improve the VLN performance. Recent approaches have begun using transformer-based architectures, such as BERT~\cite{devlin2018bert} for VLN~\cite{Hong_2021_CVPR,majumdar2020improving}. In the \emph{oracle guidance} setting, an agent may receive feedback (ground truth actions~\cite{chi2020just}, encoded ground truth action~\cite{nguyen2019vision}, or a fixed set of natural language instructions~\cite{nguyen-daume-iii-2019-help}) from an oracle during navigation. A major challenge in these works, however, is to identify when to query an oracle for feedback. In the \emph{bi-directional interaction} setting, an agent can use natural language to seek navigation help~\cite{banerjee2021robotslang, thomason2020vision, cao2022goal, lin2022multimodal}. Thomason \etal~\cite{thomason2020vision} introduced the CVDN dataset with human-human dialogue for navigation. However, these works allow the agent and oracle to communicate only at certain locations of the environment, making it less practical to real world scenarios. Self-Motivated Communication Agent (SCoA)~\cite{Zhu_2021_ICCV} permits the agent to only ask templated questions filled in with labels of detected scene objects, grossly limiting the nature of interaction between the agent and the oracle. Contrary to these methods, we empower our \name agent with: (i) the ability to seek occasional human/oracle help at any location and (ii) competence for natural language-based scene grounded conversations with an oracle for effective navigation.. Further, our agent is also robust to noisy feedback from the oracle.  

\noindent \textbf{LLM-based Embodied Navigation:} The spark of recent advancements in large language models (LLMs) ~\cite{bubeck2023sparks,touvron2023llama,openai2023gpt4} has brought along new opportunities in improving multi-modal robot navigation. In the context of Vision and Language Navigation, early works like LM-Nav~\cite{shah2022lmnav} analyzed landmarks in the instruction to be used for visual navigation. In NavGPT,~\cite{zhou2023navgpt} the possibility of integrating ChatGPT~\cite{ouyang2022training} with a vision foundation model: BLIP-2~\cite{li2023blip2} into its prompting setup to perform multi-modal reasoning to navigate in a zero-shot manner was explored. While, these works achieve decent performances on vision-language navigation task, they do not incorporate audio as part of the inputs and are thus complementary to our efforts.

\section{Proposed Method}
\textbf{Task Setup:} We assume the standard embodied audio goal problem setup ~\cite{chen2020soundspaces}, where the agent is equipped with an RGBD camera and a binaural microphone and at any time step can take one of four navigation actions: $\set{\sstop, \moveforward, \turnright, \turnleft}$ in a densely-sampled 3D grid with the goal of locating the audio source. As in~\cite{chen2020soundspaces}, we assume the sound is semantically unique and is produced by a static object, however the audio could be noisy, sporadic, or mixed up with other environmental sounds. An audio goal navigation episode is deemed successful if the agent calls the $\sstop$ action within a given proximity to the goal. 

Beyond the standard problem setup above, our CAVEN agent can also seek language-based guidance from an oracle. Practically, the oracle could be a human who has higher level information about the scene, e.g., a remote operator controlling several such agents and intervening whenever needed, or a home owner who is notified about the situation and is sought to provide guidance. To incorporate the language modality into the audio goal setup, we follow AVLEN~\cite{paul2022avlen} in which the agent can query the oracle for help and the oracle responds via a short message describing a pathlet towards the audio goal. However as is clear, the interaction in AVLEN is only uni-directional and the agent cannot ask questions. Our \name agent goes beyond this shortcoming and can phrase a question in free-form natural language using cues from the audio-visual context. Further, we assume the oracle after receiving this question, will either give a \emph{``yes''} response if the oracle's interpretation of the question in its own state space results in actions that match its estimate of the actions along the ground truth geodesic to the goal. Otherwise, the oracle responds with a \emph{``no''} followed by a short sentence guiding the agent to the goal. Note that the oracle in AVLEN has access to the 3D space of the full environment and thus can provide plausible instructions for navigation, however the \name agent has only a \emph{very restricted view} of the scene in its vicinity, thus making this task of creating a question at the agent's side entirely different from that of the oracle's. In our new problem setup, we also assume that the number of times an agent can receive direct navigation instructions from the oracle (as a result of a wrong question or when it directly queries) is limited by a budget so that the agent only seeks help when necessary.

\subsection{CAVEN Learning and Inference Framework}
\begin{figure}[t]
    \centering
  \includegraphics[width=8cm,trim={6.9cm 11.5cm 10.5cm 4cm},clip]{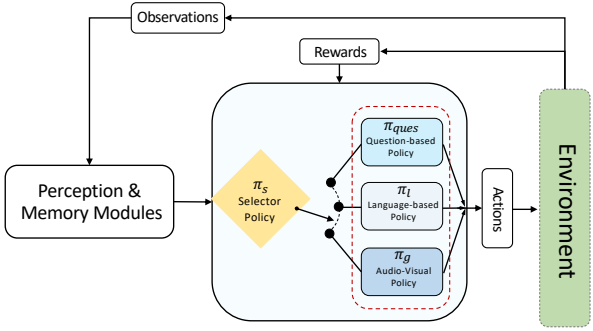}
  \caption{Architecture of our \name model. We show the reinforcement learning policies, namely a selector policy $\pi_s$ and three option policies $\pi_g,\pi_l$, and $\pi_{ques}$.}
  \label{fig:model1}
\vspace{-7mm}
\end{figure}
As we envisage \name to incorporate various modules with diverse temporal spans, it is natural to consider a partially observable semi-Markov decision process (POSMDP) as our control module 
\cite{le2018deep}. A POSMDP is essentially a partially observable Markov decision process (POMDP) with macro actions and is characterized by the tuple $(\stateset, \actionset, \transition, \Reward, \observations, \Oprobs, \discount)$ where $\stateset$, $\actionset$, $\transition$, $\Reward$, and $\discount$ are the state space, action space, transition function, reward function, and discount factor, respectively, while $\observations$ and $\Oprobs$ are the observation space and observation model. In a partially observable setup, the agent maintains a belief distribution $\Belief$ over $\stateset$, which is used to compute the expected reward. While in a POMDP setup, the agent maintains a policy $\policy:\real^{\card{\stateset}}\times \actionset \to [0,1]$ that maximizes the expected reward, in POSMDP the agent maintains multiple low level `options' as temporal abstractions, denoted $\Xi$, and a high level selector policy $\pis$ to select the options from $\Xi$. An option $\xi \in \Xi$ is defined by the triplet $(\stateset^{\xi}, \pi^{\xi}, \beta^{\xi})$, where $\stateset^{\xi}$ is the set of valid states, $\pi^{\xi}$ is the policy, $\beta^{\xi}$ is the termination condition. In our setup, we disentangle agents' interactive audio goal navigation process into three low level temporal abstractions (i.e., options): i) audio-visual navigation $\xi_g$, ii) instruction-guided navigation $\xi_l$, iii) bi-directional question-answer navigation $\xi_{ques}$. We use  $\pigoal$, $\pivln$, and $\piquestion$ to denote the respective policies of $\xi_g$, $\xi_l$, and $\xi_{ques}$ and $\Reward'_g$, $\Reward'_{l}$, and $\Reward'_{ques}$ as corresponding immediate rewards. In our case, instead of using the termination condition for each option, we allow the audio-visual navigation option $\xi_g$ to take a single step, while the interaction-based options ($\xi_l$ and $\xi_{ques}$) are allowed a fixed span of $\nu$ steps (unless $\sstop$ action is executed by these options). These options are assumed valid in any state of the environment, i.e., $\stateset_{\xi_g},\stateset_{\xi_l}, \stateset_{\xi_{ques}} \in \stateset$. 

Although the agent always has access to the three option policies, it should maintain its autonomy and should only engage in a limited number of language interactions to mitigate its uncertainty. Further, in our setup, we have different levels of engagement of the oracle with the agent for varied language interactions (e.g., bi-directional conversations with question and answer, querying for language instructions) and a system should favor asking correct questions based on its audio-visual cues over relying on oracle instructions to reduce the oracle's effort. To consider all of these scenarios, we formulate option policies with dynamically adjusted constraints. These constraints are realized by penalties associated with the reward functions of each option policy. The audio-visual navigation policy $\pigoal:\real^{\card{\stateset}\times\card{\memory}}\times \Goal \times \card{\actionset}\to [0,1]$ chooses the navigation actions $a \in \actionset$ based on the audio-visual features. Here, $M$ is a memory module storing a fixed number of past observations, and $G$ is a set of audio goal estimates. Since, $\pigoal$ is fully autonomous and does not require oracle interaction, we encourage selecting this option by defining an unconstrained reward, $\Reward'_g(\Belief_t,  \action_t)=\expect\left[\sum_{i=t}^{\infty}\discount^{i-t}\Reward_{g}'(\Belief_i, \action_i)\right]$. The instruction guided navigation policy $\pivln:\real^{\card{\stateset}\times \nu}\times \Instruction \times \Goal \times \card{\actionset}\to [0,1]$ navigates based on the received natural language instruction. Here, $\Instruction$ is the set of all natural language instructions. Since, $\pivln$ is entirely dependent on the oracle instruction, we penalize such interactions using $\zetavln$, i.e., $\Reward'_\ell(\Belief_t,  \action_t)=\expect\left[\sum_{i=t}^{t+\nu-1}\discount^{i-t}\Reward_{\ell}'(\Belief_i, \action_i)\right] - \zetavln(t)$. The bi-directional conversational navigation policy $\piquestion:\real^{\card{\stateset}\times \nu}\times \Question \times \Instruction \times \Goal \times \card{\actionset}\to [0,1]$ navigates based on asking a question and receiving an answer. Here, $\Question$ is the set of all natural language questions. Specifically, $\piquestion$ consists of multiple novel components and the policy module can be divided into three submodules based on the functionality: i) question generator $\QuesGenerator$, ii) question evaluator $\Evaluator$, and iii) instruction generator $\InsGenerator$. The output of $\piquestion$ depends on the interplay between these submodules. Question generator $\QuesGenerator$ is used to generate questions. Then, the question evaluator $\Evaluator$ evaluates on the oracle side if the question is correct. If the question is incorrect then the instruction generator $\InsGenerator$ (which mimics the oracle) generates instructions for navigation. Since, asking correct question results in minimal oracle effort in producing a response, we define a dynamic penalty based on the question by, $\Reward'_{ques}(\Belief_t,  \action_t)=\expect\left[\sum_{i=t}^{t+\nu-1}\discount^{i-t}\Reward_{ques}'(\Belief_i, \action_i)\right] - \zetaquestion(t, \Evaluator(q))$, where $q \in \Question$ and $\Evaluator(q)$ is an indicator function that checks whether the question $q$ asked by the agent falls within the range of the estimated navigation direction by the oracle, and no penalty will incur when $\Evaluator(q) = 1$.

Putting it all together, our objective to learn these policies $\pi=\set{\pis, \pigoal, \pivln, \piquestion}$ is via maximizing the value function $\Value^{\pi}(\Belief_0)$, i.e., 
\begin{align}
    \arg\max_\policy \Value^{\policy}(\Belief_0), \text{ where }\notag
\end{align}
\vspace{-0.4cm}
\begin{align}    
    & \Value^{\pi}(\Belief) = \pis(\xi_g|\Belief)\!\left[\Reward'_g +\! \sum_{\obs'\in\observations}\Oprobs'(\obs'|\Belief,\xi_g)\Value^\pi(\Belief')\right]\! \notag\\
    & +\!\pis(\xi_\ell|\Belief)\!\left[\Reward'_\ell\! + \!\sum_{\obs'\in\observations}\Oprobs(\obs'|\Belief,\xi_\ell)\Value^{\pi}(\Belief')\right]\! \notag \\
    & +\pis(\xi_{ques}|\Belief)\!\left[\Reward'_{ques} +\! \sum_{\obs'\in\observations}\Oprobs'(\obs'|\Belief,\xi_{ques})\Value^\pi(\Belief')\right]\!.
    \label{eq:posmdp}
\end{align}

\noindent Here, $\Belief'$ is the updated belief and $\Oprobs'$ is the multi-time transition function~\cite{sutton1999between} given by: $\Oprobs'(\obs'|\Belief,\xi)=\sum_{j=1}^\infty\sum_{\state'}\sum_{\state}\discount^{j}\Oprobs(\state', \obs', j|\state,\xi)\Belief(\state)$. Below, we detail the architecture of each of these policies.

\begin{figure*}[ht]
    \centering
     \includegraphics[width = 1\linewidth]{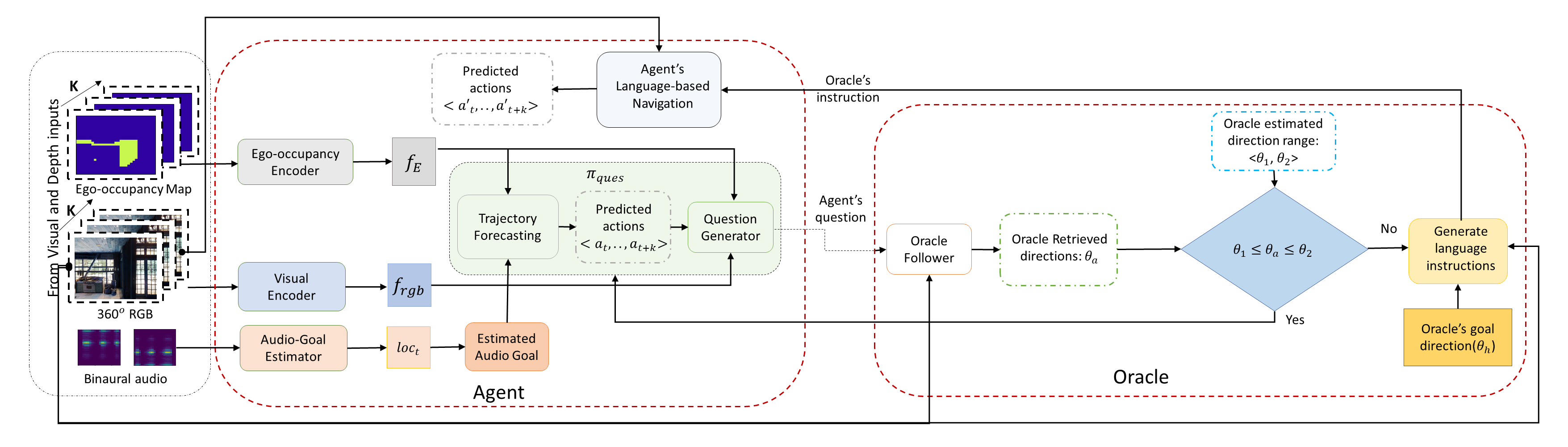} 
    \vspace{-5mm}
    \caption{Architecture of our question policy module and the control flow within it. Here, $\theta_a$ is oracle-interpreted agent's direction to take, while $\theta_1$ and $\theta_2$ represent the lower and upper bounds of oracle's estimated direction range to the goal.}
    \label{fig:ques_gen}
    \vspace{-5mm}
\end{figure*}

\noindent\textbf{Bi-directional Question-Answer Policy Module:} Bi-directional question-answer policy consists of three components: (i) \emph{TrajectoryNet} (forecasting short navigation steps), (ii) \emph{QuestionNet} (generates natural language questions using trajectories), and (iii) \emph{FollowerNet} (interprets the question on oracle side). These components detailed below are illustrated in Figure~\ref{fig:ques_gen}. They are used to enable the functionalities within the $\piquestion$ policy as: i) question generator (\emph{TrajectoryNet} + \emph{QuestionNet}), ii) question evaluator (\emph{FollowerNet}), and iii) instruction generator (\emph{QuestionNet}).

\noindent\textbf{(i)~TrajectoryNet:}
In order to forecast the steps of a trajectory, the agent needs to have a clear observation of its surroundings. Towards this end, we allow the agent to have a panoramic view at its current location. With the full view of its surroundings and an estimate of the audio-goal, the agent \emph{forecasts} a sequence of $l$-step actions, denoted by $\mathcal{F}_a$. This is achieved by \emph{TrajectoryNet} -- a transformer encoder-decoder network which takes as input a sequence of ego occupancy maps $E_t$ of four disjoint views (separated by 90-degrees) and the goal vector $\goal_t$ predicted by a binaural audio encoder, to predict a sequence of actions $\mathcal{F}_a = \langle f_{a_1}, f_{a_2} \dots, f_{a_l}\rangle$ (auto-regressively). The ego occupancy map is calculated by transforming depth images into point clouds and projecting them onto the ground plane.

\noindent\textbf{(ii) QuestionNet:}
The action sequences defined in SoundSpaces~\cite{chen2020soundspaces} are discrete, e.g,  $\moveforward$ implies moving forward by $1m$. However, the language produced from these actions by itself may be ambiguous (since it is a higher level construct) and thus does not explicitly reflect the granularity of these discrete actions. Further, as will be explained in the Experiments section, while the trajectories are forecasted using the SoundSpaces grid (which uses 90 degree angles for turning), the language instructions are produced using a model trained on the LandmarkRxR dataset~\cite{NEURIPS2021_0602940f}, that uses panoramic images as input. To compensate for these mismatches, we propose to first gather the view of the agent at the end of the forecasted trajectory, which we call $g_{view}$, and the corresponding displacement vector $g_{sub} :=[d_f, cos(\theta_f), sin(\theta_f)]$, where $d_f$ is the distance between the agent's location and the trajectory end point and $\theta_f$ is the angle difference between the direction of $d_f$ and the agent's heading direction. 

Next, we capture the panorama around the agent using 12 equiangular views, as RGB images as well as the corresponding occupancy maps to abstract the 3D scene geometry. ResNet-152 features are then extracted from these RGB images using an ImageNet pre-trained model, while the ego-occupancy maps are encoded using a 2D-CNN; both the features are fused with position embeddings and passed through a transformer encoder. In order to fuse these panoramic views with the forecasted agent views (in the SoundSpaces grid), we propose to use a transformer decoder, which takes the output of the encoder and a fusion of ResNet-152 features from $g_{view}$, coupled with the position encoding of $g_{sub}$, and the embeddings of hitherto produced words in the question (e.g., GloVe~\cite{pennington2014glove} or CLIP ~\cite{radford2021learning}), and proceeds to generate the next word in the question autoregressively.

\noindent\textbf{(iii) FollowerNet:} After the question is asked, the oracle needs to verify if it can be correctly translated into a direction that falls within the oracle's own estimation of the direction range to the goal. To this end, we incorporate \emph{FollowerNet} at the oracle, which is assumed to have knowledge of the agent's location and its audio-visual context, and can convert the question back to the oracle's space of the view angles. See Appendix for details on its  training.

\noindent\textbf{Language-based Policy Module}: There can be situations when an agent cannot produce a question to ask the oracle; e.g., when there are no useful landmarks to base the question on. To cater to such cases, we equip the agent to directly query the oracle for language-based instructions. When invoked, the agent receives instructions, similar to when a wrong question is posed to the oracle. 

\noindent\textbf{Audio-Visual Navigation Policy Module}: This policy is modeled as a transformer~\cite{vaswani2017attention} based encoder-decoder as in~\cite{chen2021semantic}. The encoder takes as input the current and previous observations in the memory $\memory$, the output of which is combined with the goal descriptor $\goal$ and decoded by the decoder to produce a feature vector defining the belief state of the agent $\Belief$. Next, a single-layer actor-critic neural network learns a policy, $\pigoal$, that transforms this belief $\Belief$ to predict the distribution on the navigation actions, which the agent samples to take a step in the environment.

\noindent \textbf{Selector Policy}:
This module, denoted $\pis$, decides when to navigate using audio-visual cues (i.e., use $\pigoal$), when to query the oracle for instructions directly (i.e., use $\pivln$); or when to pose a question to the oracle, (i.e., use $\piquestion$). Instead of directly using model uncertainty (as is common in prior works~\cite{chi2020just}), we use our proposed RL framework to train this policy in an end-to-end manner, guided by the reward design $\zeta$ described below. 

\begin{table*}[t]
\centering
  \caption{Comparison of \name performances against the state of the art under heard and unheard sound settings.}
  \vspace{-0.2cm}
  \resizebox{\linewidth}{!}{%
  \begin{tabular}{l|c|ccccccc|ccccccc}
    \toprule
    &  &\multicolumn{7}{c}{\underline{\textbf{Heard Sound}}} & \multicolumn{7}{c}{\underline{\textbf{Unheard Sound}}} \\
    & \textbf{Feedback} & \textbf{Success $\uparrow$} & \textbf{SPL $\uparrow$} & \textbf{SNA $\uparrow$} & \textbf{DTG $\downarrow$} & \textbf{SWS $\uparrow$} & \textbf{SNI $\uparrow$} & \textbf{SNO $\uparrow$} &  \textbf{Success $\uparrow$} & \textbf{SPL $\uparrow$} & \textbf{SNA $\uparrow$} & \textbf{DTG $\downarrow$} & \textbf{SWS $\uparrow$} & \textbf{SNI $\uparrow$} & \textbf{SNO $\uparrow$} \\    
    \hline   
    
    Random  Nav. & \xmark & 1.4 & 3.5 & 1.2 & 17.0 & 1.4 & - & - & 1.4 & 3.5 & 1.2 & 17.0 & 1.4 & - & -\\
    ObjectGoal RL & \xmark & 1.5 & 0.8 & 0.6 & 16.7 & 1.1 & - & - & 1.5 & 0.8 & 0.6 & 16.7 & 1.1 & - & - \\
    Gan et al. \cite{gan2020look} & \xmark & 29.3 & 23.7 & 23.0 & 11.3 & 14.4 & - & - & 15.9 & 12.3 & 11.6 & 12.7 & 8.0 & - & - \\
    Chen et al. \cite{chen2020soundspaces} & \xmark & 21.6 & 15.1 & 12.1 & 11.2 & 10.7 & - & - & 18.0 & 13.4 & 12.9 & 12.9 & 6.9 & - & - \\
    AV-WaN \cite{chen2021learning} & \xmark & 20.9 & 16.8 & 16.2 & 10.3 & 8.3 & - & - & 17.2 & 13.2 & 12.7 & 11.0 & 6.9 & - & -\\
    SMT\cite{fang2019scene}+Audio  & \xmark & 22.0 & 16.8 & 16.0 & 12.4 & 8.7 & - & - & 16.7 & 11.9 & 10.0 & 12.1 & 8.5 & - & - \\
    SAVi (Chen et al., 2021) & \xmark & 33.9 & 24.0 & 18.3 & 8.8 & 21.5 & - & - & 24.8 & 17.2 & 13.2 & 9.9 & 14.7 & - & - \\
   
    AVLEN (Paul \etal, 2022) & Language & 36.1 & 24.6 & 19.7 & 8.5 & 23.1 & - & 21.8 & 26.2 & 17.6 & 14.2 & 9.2 & 15.8 & - & 15.9\\
    AVLEN (Paul \etal, 2022)  & GT Actions & 48.2 & 34.3 & 26.7 & 7.5 & 36.0 & - & 29.1 &  36.7 & 24.1 & 18.7 & 8.3 & 26.6 & - & 22.3\\
     \hline
    \textbf{\name (Ours)}  & Noisy-Language & \textbf{45.2} & \textbf{32.9} & \textbf{28.8} & \textbf{7.5} & \textbf{32.3} & 17.9 & \textbf{31.4} & \textbf{38.2} & \textbf{27.6} & \textbf{24.1} & \textbf{8.2} & \textbf{25.9} & 15.0 & \textbf{23.1}\\
    \textbf{\name (Ours)}   & Language & \textbf{48.4} & \textbf{35.8} & \textbf{31.0} & \textbf{6.9} & \textbf{34.2} & 21.5 & \textbf{33.4} & \textbf{42.0} & \textbf{30.0} & \textbf{26.5} & \textbf{7.6} & \textbf{30.9} & 16.7 & \textbf{27.9}\\
    \textbf{\name (Ours)} & GT Actions & \textbf{54.8} & \textbf{41.4} & \textbf{35.9} & \textbf{6.5} & \textbf{39.9} & 24.3 & \textbf{37.8} & \textbf{49.7} & \textbf{37.3} & \textbf{32.7} & \textbf{6.7} & \textbf{37.2} & 19.8 & \textbf{33.0} \\

    \bottomrule
  \end{tabular}%
}
  \label{tab:heard_unheard_main}
\vspace{-0.3cm}
\end{table*}

\subsection{Reward Design}
In this section, we detail the rewards structure to train the various policy modules in an end-to-end manner. For the $\pigoal$ policy, we use the reward scheme in~\cite{chen2020soundspaces}, i.e., the agent gets $+1$ for moving towards the goal and receives $+10$ if it calls the $\sstop$ near the goal. Further, to make the navigation efficient, we penalize by $-0.01$ for every step taken. The  penalty structure for the language-based policies is designed so as to discourage the agent to seek help from the oracle, while also limiting the number of instructions $K\geq 0$ received. To this end, we propose a dynamic penalty that increases in magnitude as more instructions are sought from the oracle. Specifically, if $\zeta_l(k, K)$ denotes the penalty received by the agent for the $k$-th query, then
\begin{align}
\zeta_l(k, K) = \begin{cases} 
      \frac{k \times (\reward_{neg} + \exp({-\nu}))}{\nu} & k< K \\
      \reward_{neg} + \exp({-k})  & k\geq K, 
   \end{cases} 
\end{align}
where $\nu$ characterizes the number of steps the agent takes based on the language instruction received, which is fixed in our case, and $\reward_{neg}=-0.6$ is a constant. Until $k<K$, the penalty is linear, however for $k\geq K$, the penalty approaches $\reward_{neg}$ exponentially thereby discouraging the agent to seek language guidance directly. Further to this penalty, we also include an additional cost for seeking oracle guidance frequently. Specifically, we include a linear penalty $\zeta_f$ if the agent queries the oracle within $\tau$ steps, where $\zeta_f(j)=  \frac{\reward_{f}}{j}$ for the $j$-th step, if $j\in[0,\tau]$ and zero otherwise (with $\reward_f=-0.5$). Thus, the total penalty for the agent querying the oracle is given by $\zeta_l + \zeta_f$. 

As the question policy $\piquestion$ blends between $\pigoal$ and $\pivln$, we propose a penalty structure that integrates both these policies. Specifically, if $\zeta_{ques}(m)$ is the penalty incurred by the agent for asking the $m$-th question, then 
\begin{equation}
\zeta_{ques}(m) = \zeta_l(m, K')\ \delta_{ques}(m) + \zeta_{f_{ques}}(m),
\end{equation}
where $\zeta_{f_{ques}}$ is the penalty for asking questions too many times (similar to $\zeta_f(k)$), $K'$ is the budget on the number of wrong questions, and $\delta_{ques}(m)=1$ if the response to the $m$-th question by the oracle is ‘no’, else $\delta_{ques}(m)\in [0,1)$ is a constant. In our experiments, we find that not penalizing the agent for correct questions leads to better results, i.e., $\delta_{ques}(m)=0$. Such a differential reward  implicitly reinforces the agent to learn correct trajectories to the audio goal, improving performance. We also couple $\piquestion$ with $\pivln$ by enforcing $K + K'=\eta$ for an $\eta=3$. Using this reward setup, the policies are trained with the DD-PPO algorithm~\cite{wijmans2019dd}. See the Appendix for details on policy training.

\section{Experiments}
\label{sec:expts}
\noindent \textbf{Datasets:} The \name agent is trained and evaluated on the SoundSpaces platform~\cite{chen2020soundspaces}. It uses MatterPort3D environment scans~\cite{chang2017matterport3d}. We use the the semantic audio-visual navigation dataset from ~\cite{chen2020soundspaces} to benchmark our experiments. The details of the dataset are provided in the Appendix.

\noindent\textbf{AVN-Instruct Dataset:}
For pre-training and evaluation of the language interaction modules (i.e., \emph{QuestionNet}, \emph{FollowerNet}), we use the Landmark RxR dataset~\cite{NEURIPS2021_0602940f}, which contains 150k well-annotated sub-trajectories and their corresponding language sub-instructions grounded on scenes captured using the MatterPort3D simulator. Then we adopt the pre-trained QuestionNet to synthesize a dataset called AVN-Instruct, which contains a total of 41.5k dense pairs of sub-instructions, audio-goal, and visual scene under the state space of the Soundspace Habitat simulator, by sampling the trajectories and transporting the grid from Matterport3D to Soundspace and obtaining the action sequence which closely approximates this trajectory. Before integrating the modules into the RL framework, we fine-tune the whole question module end-to-end on AVN-Instruct with a set of 500 and 1000 samples for validation and testing. 

\noindent \textbf{Evaluation Metrics:} We follow the standard metrics defined in SAVi~\cite{chen2021semantic} to evaluate the navigation performance, namely: (i) success rate (SR) for navigation success, (ii) success rate weighted by inverse path length (SPL), (iii) success rate weighted by inverse number of actions (SNA), (iv) average distance to goal (DTG), and (v) success rate when silent (SWS). In addition, we introduce two new metrics for assessing navigation performance that also considers the number of language-based oracle interactions, namely: (a) success rate weighted by the inverse number of language interactions (\underline{SNI}) -- which is the ratio of the success rate to the average total number of times either direct instructions are sought from the oracle or a question is posed to it (averaged by the number of episodes), and (b) success rate weighted by inverse number of oracle instructions (\underline{SNO}) -- which is the ratio of the success rate to the average total number of times either direct instructions are sought from the oracle or a \textit{wrong} question is posed to it. These additional metrics help explain the performance gain under conversational settings.

\begin{table}[t]
    \centering
    \caption{ Comparison of \name performances 
    with different approaches in the \emph{presence of distractor sound.}}
    \vspace{-0.2cm}
    \resizebox{\linewidth}{!}{
    \begin{tabular}{l|c|@{\hskip 3mm}c@{\hskip 3mm}c@{\hskip 3mm}c@{\hskip 3mm}c@{\hskip 3mm}c@{\hskip 3mm}c@{\hskip 3mm}c}
    \toprule
    & \textbf{Feedback} & \textbf{Success $\uparrow$} & \textbf{SPL $\uparrow$} & \textbf{SNA $\uparrow$} & \textbf{DTG $\downarrow$} & \textbf{SWS $\uparrow$} & \textbf{SNI $\uparrow$} & \textbf{SNO $\uparrow$} \\ \hline   
    Chen et al  & \xmark & 4.0 & 2.4 & 2.0 & 14.7 & 2.3 & - & - \\
    AV-WaN \cite{chen2021learning} & \xmark & 3.0 & 2.0 & 1.8 & 14.0 & 1.6 & - & - \\
    SMT+Audio \cite{fang2019scene}  & \xmark & 4.2 & 2.9 & 2.1 & 14.9 & 2.8 & - & -\\
    SAVi (Chen \etal, 2021) & \xmark & 11.8 & 7.4 & 5.0 & 13.1 & 8.4 & - & - \\[1mm]
    AVLEN (Paul \etal, 2022) & Language  &  14.0 & 8.4 & 5.9 & 12.8 & 11.1 & - & 8.5\\
    \hline
    Random & Bi-interact & 16.9 & 10.6 & 7.9 & 11.9 & 11.1 & 7.2 & 9.4 \\
    Uniform & Bi-interact & 16.9 & 10.5 & 7.6 & 11.9 & 11.6 & 7.1 & 9.5 \\
    Model Uncertainty  & Bi-interact & 19.6 & 12.4 & 8.9 & 11.4 & 14.0 & 7.8 & 10.2 \\[1mm]
    \hline
    \name & Bi-interact  &  \textbf{21.3} & \textbf{13.9} & \textbf{11.7} & \textbf{11.6} & \textbf{14.5} & 8.4 & \textbf{11.6} \\ \hline
    \end{tabular}
    }
    \label{tab:distractor_main}
    \vspace{-0.4cm}
\end{table}

\begin{table}[t]
\centering
    \caption{Ablation of the reward parameter $\delta_{ques}$ of \name's question module under unheard sound settings.} 
    \resizebox{1\linewidth}{!}{
    \begin{tabular}{l|ccccc}
    \toprule
    \textbf{Architecture} & \textbf{Success $\uparrow$} & \textbf{SPL $\uparrow$} & \textbf{SNA $\uparrow$} & \textbf{DTG $\downarrow$} & \textbf{SWS $\uparrow$} \\    \hline
    \name ($\delta_{ques}$=1.0) & 32.1 & 23.1 & 19.4 & 8.0 & 20.8 \\
    \name ($\delta_{ques}$=0.5) & 36.5 & 26.9 & 24.6 & 8.2 & 21.1 \\
    \name ($\delta_{ques}$=0.0) (ours) & \textbf{42.0} & \textbf{30.0} & \textbf{26.5} & \textbf{7.6} & \textbf{30.9} \\
    \bottomrule
    \end{tabular}
    }
    \label{tab:reward_parameter_ablation}
    \vspace{-0.4cm}
\end{table}

\noindent \textbf{Experimental Results and Analysis:}   Here, we compare our proposed formulation against state-of-the-art semantic audio-visual navigation
approaches, namely~\cite{gan2020look},~\cite{chen2020soundspaces}, AV-WaN (Chen \etal, 2021), SMT~\cite{fang2019scene} + Audio, SAVi (Chen \etal, 2021) and AVLEN~(Paul \etal, 2022). Using the same protocol as in AVLEN, we evaluate our performances on two different settings: (i) heard and (ii) unheard sound, both in unseen environments with sporadic sources. To ensure the comparisons are fair, we control our \name model to have a similar number of oracle feedbacks as  in Paul \etal. Table~\ref{tab:heard_unheard_main} provides the results of our experiments using heard and unheard sounds. The table shows that our full model --\name (language), is capable of achieving significant improvements across all metrics. \name exhibits a \textbf{12\% gain} on the newly introduced \textbf{SNO} metric over Paul \etal, our closest competitor, in both heard and unheard cases. This clearly shows that the agent benefits much more from both our novel language components. Given the budget on directly receiving instructions from the oracle, we find that \name poses a correct question about $40\%$ of the time, thereby incurring less penalty. Even with a noisy oracle, i.e., \emph{Noisy-Language} in Table~\ref{tab:heard_unheard_main}, we achieve better performances compared to Paul \etal, showing the robustness of our framework. To induce noise, we either ground the generated oracle's instructions on random trajectories or switch 'yes'/'no' responses, both with a chance of $25$\%.

\noindent \textbf{Navigation Under Distractor Sounds:} We also evaluate the performance of \name in the presence of distractor sounds, in the unheard setting. Since this environment presents a mixture of sounds, therefore to disambiguate, a one hot encoding of the target sounding object is also provided as an input to the agent (as is the standard evaluation protocol~\cite{chen2021semantic}).  The presence of distractor sounds adversely affects the estimation of the audio-goal, which results in more uncertainty in the agent's decision-making. Under this setting, the conversations between the agent and oracle becomes even more critical. Even under such challenging circumstances, as shown in Table~\ref{tab:distractor_main}, we notice a \textbf{$5.5\%$} and a \textbf{$3.1\%$ gain} on SPL and SNO, respectively against our closest competitor. 

\noindent\textbf{Ablation on Selector Policy:} In Tables~\ref{tab:heard_unheard_main},~\ref{tab:distractor_main}, we compare various strategies instead of learning the selector policy, $\pi_s$. In \emph{Random}, the agent randomly selects a navigation policy, while in \emph{Uniform}, the agent chooses a policy every 3 steps, alternating between the three policies. In Model Uncertainty, the audio-goal uncertainty estimated by the selector policy is used to decide which policy to invoke; i.e., if the audio-goal uncertainty is above 66.7\%, the language-based policy is invoked; if the uncertainty is between 33.3\% and 66.7\%, question policy is invoked; otherwise, the audio-goal policy is invoked. Our results show learning of $\pi_s$ is better.

\begin{figure}[!htbp]
\centering
  \includegraphics[trim={0.2cm 2cm 0.3cm 2cm}, clip,width=\linewidth]{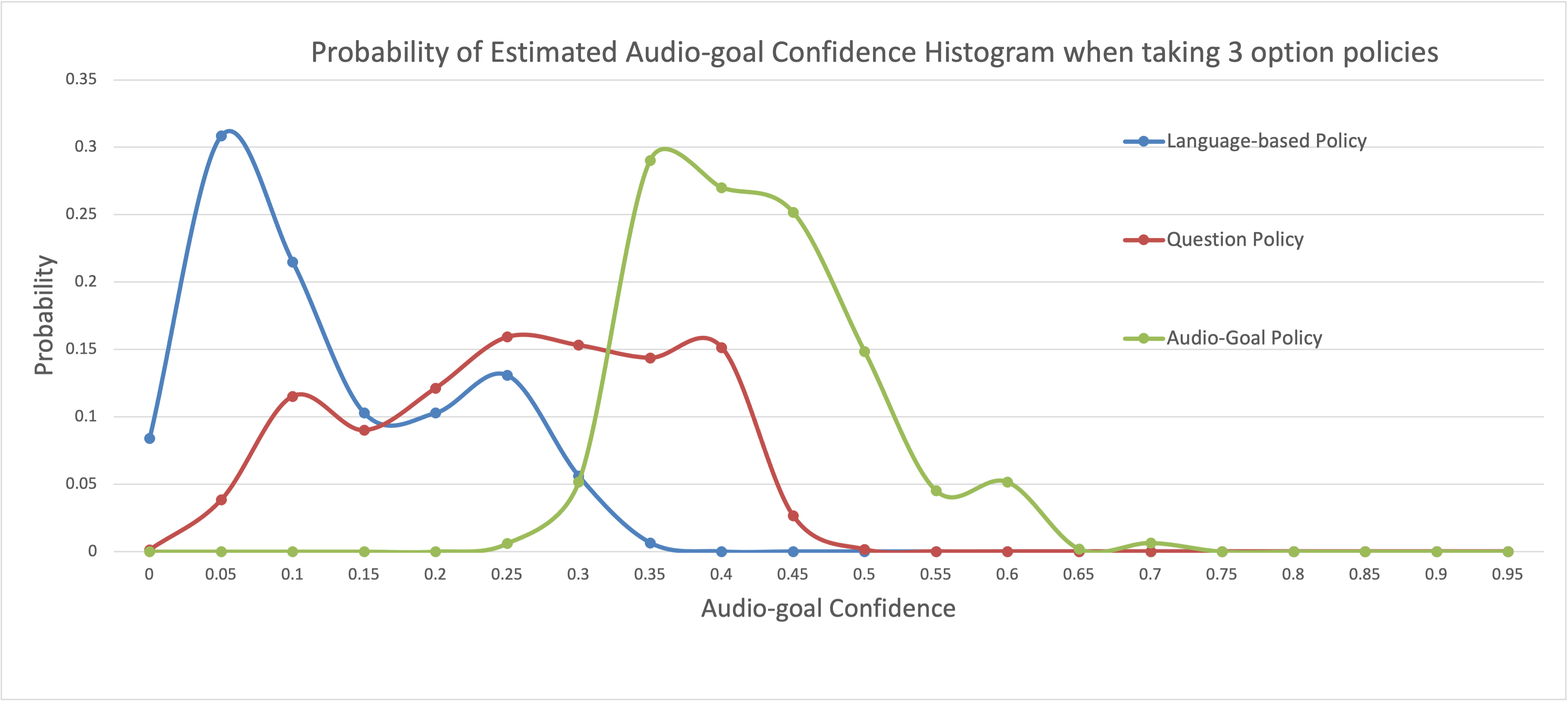}
  \caption{Distribution of estimated audio goal confidence when each policy is invoked.}
  \label{fig:gdist}
  \vspace{-2mm}
\end{figure}

\noindent \textbf{Analysis of Policy Dynamics:} To study the situations when the agent invokes the various navigation policies, we record the confidence of the audio-goal estimated by the selector policy $\pi_s$, when each of the option policies is invoked and compute its distribution using all test set episodes. As shown in Figure~\ref{fig:gdist}, the audio-goal is invoked when the agent is highly confident and the language-based policy is invoked when the agent's confidence is low. It is note-worthy that the question policy is invoked more often when the agent is moderately confident. Though it potentially risks being penalized by asking wrong questions, it benefits from seeking confirmation from the oracle using its own audio-visual cues to help alleviate navigation uncertainty, thus facilitating efficient navigation.

\noindent\textbf{Insights into Differential Rewarding:}  In Table~\ref{tab:reward_parameter_ablation}, we report the \name performances on varying the penalty parameter $\delta_{ques}$. Note that our differential rewarding scheme gives no penalty when the agent makes a correct question $\delta_{ques}=0$, however penalizes heavily for mistakes. Thus, the \emph{gap} between the two penalties acts as an incentive for the agent to make more number of correct trajectory predictions than in a case where this penalty gap is lower (e.g., $\delta_{ques}=0.5, 1.0$ in which case it is similar to the penalty it receives for the wrong question). The success rate is much higher suggesting that the incentive the agent receives in making a correct question influences the learning of the trajectory forecasting significantly more.
 
\section{Conclusions}
In this paper, we introduced \name for embodied navigation in an audio-visual setting for the audio goal navigation task, where the agent is also equipped to converse with an oracle in natural language, when uncertain. We introduced a novel budget-aware partially observable semi-Markov decision process to learn the various control policies for solving the task. Quantitative evaluations of \name under various noisy problem settings, using established and novel metrics, demonstrate large improvements in performance over competing methods, substantiating the benefits of our proposed interaction policies and our architecture. However, the interactions with the oracle might result in the agent having to wait for feedback, which we intend to fix in future work. 

\appendix
\section{Appendix}
\subsection{A1. Details of the SoundSpaces dataset}

The \name agent is trained and evaluated on the SoundSpaces platform~\cite{chen2020soundspaces} to perform audio-visual-language navigation. It uses Matterport3D environment scans~\cite{chang2017matterport3d}, and creates a realistic simulation of densely sampled grid of the 3D space with $1m$ resolution driven by the Habitat-Simulator engine ~\cite{habitat19iccv}. The simulation platform provides panoramic egocentric views of the agent in terms of both RGB and depth images and also provides binaural acoustics in 3D space. We use the the semantic audio-visual navigation dataset from Chen \etal~\cite{chen2020soundspaces} to benchmark our experiments. The dataset consists of 21 semantic categories of sound associated with different objects in the Matterport3D scans. The duration of the sounds follow a Gaussian distribution with mean 15s and a standard deviation of 9s. There are 500,000/500/1000 episodes in the train/val/test splits, respectively, derived from 85 scans. For navigation experiments, we evaluate on 1000 episodes from the test split using 3 different evaluation protocols defined in semantic audio-visual navigation~\cite{chen2021semantic} task, respectively `heard', `unheard' and `distractor sound' (representing noisy audio-goal scenarios).

\subsection{A2: Details of the AVN-Instruct dataset}
\label{sec:avn}

\begin{figure*}[t]
\begin{center}
  \includegraphics[width = \linewidth]{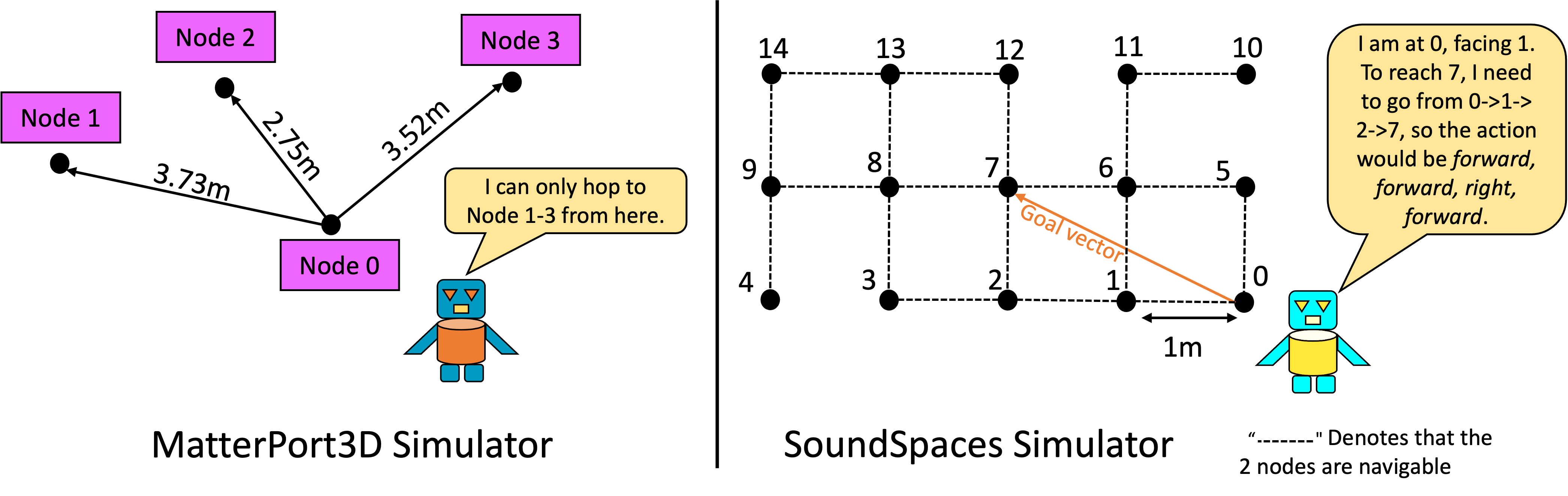}
  \caption{We compare the grid sturctues of MatterPort3D (Left) with that of SoundSpaces simulator (Right) side by side. There are 2 main differences: i). The MatterPort3D simulator has an \textbf{irregular} grid structure while the SoundSpaces simulator has a regular square grid. ii). Average distance between adjacent nodes in the MatterPort3D is larger than SoundSpaces simulator (1m apart).}
  \label{fig:sim}
  \end{center}
\vspace{-0.7cm}
\end{figure*}

In order to equip our \name agent to reason using audio, visual, and language inputs, language sub-instructions need to be introduced into the SoundSpaces simulation environment. Existing vision-and-language sub-instruction datasets, such as the Landmark RxR dataset, are based on the Matterport3D simulator, which has a different grid setup and action space from that of SoundSpaces. Thus, even though SoundSpaces simulator uses  MatterPort data scans, it results in a very dense 3D grid structure and can render dense observations compared to MatterPort3D simulator, which has a sparse, irregular grid. Specifically, the Matterport3D grid points are around 3 meters apart on average compared to that in the dense grids of SoundSpaces which are precisely 1m apart with a regular square arrangement. In Figure~\ref{fig:sim}, we present the differences between the grid structures of MatterPort3D and SoundSpaces by means of a graphical example for the reader's convenience.

To bridge this gap between the Landmark RxR sub-instructions available on the Matterport3D grid and SoundSpaces, a first approach would make direct grid approximations by overlaying the Matterport3D grid on top of the SoundSpaces grid. In such a scenario, for each grid point with a labeled language sub-instruction on Matterport3D (data samples in Landmark-RxR), we could find the closest grid point in the SoundSpaces environment, then map the respective trajectory from the Matterport3D dataset into SoundSpaces  with the associated language instruction. However, we found that such approximations incur a large error, since the Matterport3D grid is much more sparse than its SoundSpaces counterpart (1m apart for adjacent grid points). Even imposing reasonable thresholds -- e.g.,  discarding any trajectory with grid approximation error larger than 0.25m -- results in over 98\% of the original instruction trajectories from Landmark-RxR dataset being discarded. This in turn leads to a very limited number ($\sim$3k) of trajectories associated with sub-instructions (recall that the sub-instructions with their associated trajectories form the training set for QuestionNet to produce its questions or oracle instructions, and FollowerNet to decode goal directions). Such a limitation prevents the language interaction of the question module, i.e., QuestionNet and FollowerNet, from being effectively trained and fine-tuned on the SoundSpaces environment.

Towards this end, we use another approach by applying our pretrained navigation language model, i.e., QuestionNet (bootstrapped via training on the data from the first approach) to generate tuples of (audio goal, trajectory, sub-instructions) directly on the SoundSpaces environment via a self-supervised approach. That is, we randomly sample audio goals and their trajectories from SoundSpaces existing training episodes, which are then mapped to goal vectors and passed to the pre-trained QuestionNet to produce language sub-instructions directly on the SoundSpaces grid. These sub-instructions are then passed as input to the FollowerNet, which needs to decode them to the goal directions and the trajectories. The entire pipeline of QuestionNet and FollowerNet is trained end-to-end so that both modules improve. We use a small learning rate for QuestionNet which aims to prevent the model from drifting away from the pre-trained model and produce valid natural language instructions as our empirical results substantiate.  

The above approach, thus could generate sub-instructions by grounding the QuestionNet on any navigable sub-trajectories in the SoundSpaces dataset. Further, the intermediate sub-instructions generated by the approach could be used as a new dataset for mapping SoundSpaces actions to sub-instructions. Motivated by this observation, we synthesized an additional 38.5k sub-instructions data. Combining it with the 3k sub-instructions by grid approximation (1st approach), we created a large-scale audio-visal-language sub-instruction dataset -- AVN-Instruct, where each sample consists of a triplet of audio-goal vector, oracle action sequence (defined in action space of SoundSpaces environment), and language sub-instructions. AVN-Instruct consists of a total of 41.5k dense pairs of sub-instructions. We will be releasing the pretrained model - QuestionNet for generating instructions, the AVN-Instruct dataset, and the code for performing grid approximations and synthesizing the instructions. In addition, this particular approach for introducing language instructions into audio-visual embodied platform is not described in previous literature, and thus is valuable to provide a useful guide for researchers who are interested in tri-modal navigation. 

\subsection{A3: Qualitative Results}
\label{sec:quals}
\begin{figure}[t]
\begin{center}
  \includegraphics[width = \linewidth]{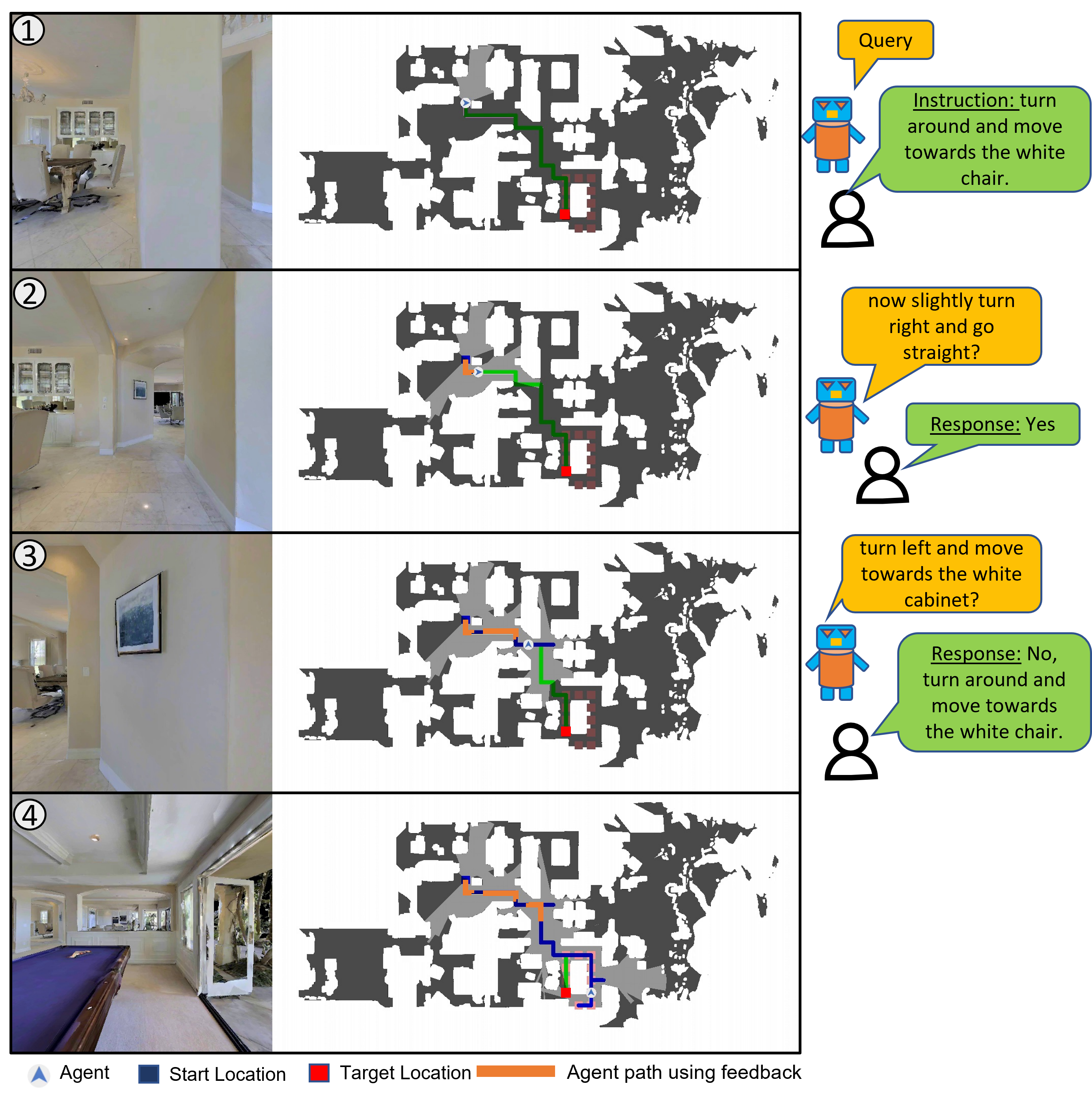}
  \caption{We show egocentric views and top-down maps for four different viewpoints in the agent's trajectory. The agent starts from  ~\textcircled{1} where it queries for help and receives instruction, in \textcircled{2} asks a question and receives a positive response, in \textcircled{3} asks a question based on an incorrect predicted trajectory hence receives correct trajectory instructions, and finally reaches near the goal in \textcircled{4}.}
  \label{fig:qualitative}
  \end{center}
\vspace{-0.7cm}
\end{figure}

We show a qualitative example of navigation using \name in Figure~\ref{fig:qualitative} illustrating snapshots from an example episode. The sounding object is a ‘table’ in this example. During navigation, the agent queries for help in \textcircled{1}, asks questions in \textcircled{2} and \textcircled{3}, and finally reaches near the Audio-Goal in \textcircled{4}. 

\subsection{A4: Additional Ablation Experiments}
\label{sec:ablations}
In this section, we present ablation experiments on the various components of our \name framework. 

 \noindent \textbf{Ablations on QuestionNet}
To prove the capability of pretrained navigation language model on the 150k Landmark-RxR sub-instructions dataset (in MatterPort3D simulator) on our QuestionNet, we evaluate our generated questions or instructions using metrics commonly used for assessing the quality of language translation or image caption generation, including BLEU scores (BLEU-4)~\cite{papineni2002bleu}, METEOR score~\cite{banerjee2005meteor}, ROUGE-L~\cite{lin2004rouge}, and CIDEr~\cite{vedantam2015cider} scores against the reference instructions under the same navigation sub-trajectories. We attempted multiple variants of input encodings and as shown in Table \ref{tab:question_generation_module_ablation}, the transformer setup with CLIP embedding outperforms the Gated Recurrent Unit (GRU) or GloVe embedding-based variants. Such a strong performance on the metrics suggests that QuestionNet is capable of generating high quality sub-instructions for oracle or asking reasonable questions for the agent.

\begin{table}[t]
    \caption{Language generation performances for various architectural choices.}
    \resizebox{\linewidth}{!}{
    \begin{tabular}{l|cccc}
    \toprule
    & \textbf{BLEU-4} & \textbf{METEOR} & \textbf{ROUGE-L} & \textbf{CIDEr} \\    \hline
    GRU (GloVe embedding) &   0.301 & 0.195 & 0.537 & 1.542 \\
    Transformer (GloVe embedding) & 0.341 & 0.232 & 0.585 & 1.813 \\
    Transformer (CLIP embedding) &  0.367 & 0.263 & 0.617 & 1.942 \\
    \bottomrule
    \end{tabular}
    }  
    \label{tab:question_generation_module_ablation}
\end{table}
 
 \noindent \textbf{Ablations on Trajectory Forecasting module:}
The trajectory forecasting module takes visual observations of 360 degree views along with the predicted audio-goal vector as input, and forecasts next 4-step actions to take. Such visual observations could be RGB, RGBD, or egocentric occupancy maps. Egocentric occupancy maps provide a compact while useful representation of the obstacles around the agent. To assess the effectiveness of using egocentric occupancy maps as the sensory input to trajectory forecasting module, we evaluate the accuracy of  actions over four steps against first 4 action steps of the shortest path in terms of geodesic distance to the audio-goal. We compare our visual observation input against other input choices (RGB, RGB-D), and observe that the egocentric occupancy map inputs yield the best action forecasting accuracy (see Table~\ref{tab:trajectory_forecasting_ablation}). It is noteworthy that our trajectory forecasting network only applies a compact 3-layer CNN to encode the features of the egocentric occupancy maps, and outperforms the model that takes a pretrained ResNet-18 on ImageNet to encode RGB or RGB-D visual observations, which signifies the choice of our input modality.

\begin{table*}[t]
    \caption{ Comparison of the effectiveness of different sensor input representations for the trajectory forecasting module.}
    \centering
    \begin{tabular}{l|ccccccc}
    \toprule
    \textbf{Inputs} & \textbf{1 step acc} & \textbf{2 steps acc} & \textbf{3 steps acc} & \textbf{4 steps acc}\\    \hline
    RGB & 75.8\% & 55.3\% & 44.8\% &  38.7\%  \\
    RGB+D & 76.7\% &  66.1\%  & 57.3\%  & 50.3\% \\
    Egocentric Occupancy Map & 85.2\% & 74.9\% & 65.9\% & 58.1\% \\
    \bottomrule
    \end{tabular}
    \label{tab:trajectory_forecasting_ablation}
\end{table*}

\noindent \textbf{Ablations on language-based policy module}
To ablate our agent's language-based policy module, we replace our instruction generation (QuestionNet) and follower module (FollowerNet) with Paul \etal's language module, while keeping the other parts of the system (i.e., $\pi_g$, $\pi_s$, etc.) the same.  Paul \etal's language module adopts a simple LSTM based model that takes the continuous trajectory path as an input to generate questions following ~\cite{fried2018speaker}, such model architecture: i) is perhaps weak in model capacity as only an LSTM is applied to generate the language, and ii) only works when the agent/oracle moves along a certain trajectory, which is especially impractical for agent since actual action steps need be taken to just ask a question. Our proposed QuestionNet equips the transformer architecture with delicate visual and goal groundings, and the agent or oracle could ask questions/give instructions without taking actions.  We observe that our language-based policy  module achieves a $4.7\%$ gain on navigation success rate and a $2.4\%$ gain on SPL when both our model and Paul etal. are  pre-trained on the same Landmark RxR dataset (see Table~\ref{tab:vln_module_ablation}). 

\begin{table*}[t]
    \caption{Comparison of language-based policy module performance versus other alternatives.} 
    \centering
    \begin{tabular}{l|ccccccc}
    \toprule
    \textbf{Instruction generation and follower setup} & \textbf{Success $\uparrow$} & \textbf{SPL $\uparrow$} & \textbf{SNA $\uparrow$} & \textbf{DTG $\downarrow$} & \textbf{SWS $\uparrow$} \\    \hline
    Paul \etal's module (Fine-Grained R2R) & 26.2 & 17.6 & 14.2 & 9.2 & 15.8  \\
    Paul \etal's module (Landmark RxR) & 26.4 & 18.2 & 14.4 & 8.9 & 16.0 \\
    Ours (Landmark RxR) & \textbf{31.1} & \textbf{20.6} & \textbf{16.4} & \textbf{8.6} & \textbf{21.3} \\
    \bottomrule
    \end{tabular}
    \label{tab:vln_module_ablation}
\end{table*}

\noindent \textbf{Comparison of Different Query Triggering Alternatives:}
Table~\ref{tab:heard_unheard_query} compares the agent's performance under different query triggering methods. The alternatives considered are: (i) \textit{Uniform}: Queries are posed once every 3 steps, (ii) \textit{Random}: Actions are taken randomly in each episode for the first 50 steps, and (iii) \textit{Model Uncertainty (MU)} based on Chi \etal~\cite{chi2020just}: Queries are posed when the difference between the top-2 output action probabilities of $\pigoal$ is less than a threshold ($\leq 0.1$). For all these three scenarios, whenever a query is posed, we randomly sample whether to select the question-answer policy module or the language-based policy module according to the ratio of interactions with the question-answer branch with respect to the language-based policy branch, in the \name framework 
 (so as to make them comparable). 
The results reveal that our formulation, trained with reinforcement learning, outperforms other competing variants by a large margin, viz. \textbf{6\% on SR}, \textbf{7\% on SWS}, and \textbf{3\% on SNO}. Note that, more specifically, our model better (implicitly) captures the uncertainty of the agent than the model uncertainty heuristically estimated from the output of goal policy. 

\begin{table*}[!ht]
\centering
  \caption{Comparison between different bi-directional interaction triggering methods under heard and unheard sound settings.}
  \resizebox{\linewidth}{!}{
  \begin{tabular}{l|c|ccccccc|ccccccc}
    \toprule
    &  & \multicolumn{7}{c|}{\underline{\textbf{Heard Sound}}} & \multicolumn{7}{c}{\underline{\textbf{Unheard Sound}}} \\
    & Feedback & Success $\uparrow$ & SPL $\uparrow$ & SNA $\uparrow$ & DTG $\downarrow$ & SWS $\uparrow$ & SNI $\uparrow$ & SNO $\uparrow$ & Success $\uparrow$ & SPL $\uparrow$ & SNA $\uparrow$ & DTG $\downarrow$ & SWS $\uparrow$ & SNI $\uparrow$ & SNO $\uparrow$\\    
    \hline   
    
    Random & Language & 37.8 & 27.5 & 22.9 & 7.7 & 22.1 & 18.6 & 28.9 & 27.9 & 19.8 & 16.5 & 8.7 & 16.0 & 12.7 & 20.9\\
    Uniform  & Language & 38.9 & 27.8 & 22.8 & 7.5 & 24.1 & 18.9 & 28.7 & 26.4 & 18.5 & 15.7 & 8.4 & 14.2 & 12.6  & 20.8  \\
    Model Uncertainty & Language & 42.4 & 30.6 & 25.2 & 7.4 & 27.7 & 20.6 & 30.2 & 33.3 & 23.2 & 18.6 & 7.8 & 21.6 & 14.7 & 22.7 \\
    \textbf{\name (Ours)}  & Language & \textbf{48.4} & \textbf{35.8} & \textbf{31.0} & \textbf{6.9} & \textbf{34.2} & \textbf{21.5} & \textbf{33.4} & \textbf{42.0} & \textbf{30.0} & \textbf{26.5} & \textbf{7.6} & \textbf{30.9} & \textbf{16.7} & \textbf{27.9} \\
    \bottomrule
  \end{tabular}
  }  \label{tab:heard_unheard_query}
\end{table*}

\noindent \textbf{Ablations on Number of Branches:}
Table~\ref{tab:branch_ablation} shows the \name agent's performance under different branching setups. In particular, we compare our 3-branch setup (goal policy, language-based policy module and question-answer policy module) versus several other plausible branch setups. \name + language-based policy module is a 2-branch setup where only the audio-visual navigation policy module and the language-based policy module can be invoked whenever the agent queries. \name + question branch with/without the oracle instruction branch allows for the selection of only the audio-visual navigation policy module and the question-answer policy module. In this scenario, whenever the agent asks a question, the oracle responds with a "Yes/No" by comparing the estimated direction with the output of the \emph{FollowerNet}. The oracle can then decide \textbf{whether or not to provide oracle instructions in natural language}. If it does, then we call the model \name + question branch with oracle instructions (2-branch), otherwise \name + question branch without oracle instruction (2-branch). In the latter setting, the agent will take the audio-visual navigation policy action as the next step, if the question is a mismatched one. Such a setting simulates the scenario of a \textit{weak} feedback oracle where the agent can enhance its belief by receiving `yes' but cannot correct its belief when receiving `no'.  In \name (ours) + GT actions, we use a 3-branch setup but here the oracle gives the ground truth action as feedback instead of the oracle language instructions. As shown in the table, our proposed 3-branch setup outperforms all other 2-branch setups by a large margin. In particular we notice gains of \textbf{5.3\% SR} and \textbf{3.4\% SPL} over \name + question branch with oracle instructions. Moreover, our model's performance is reasonably close to the setting where GT actions are provided as feedback underscoring the efficacy of our model.

\begin{table*}[!ht]
\centering
    \caption{Ablation of the 3-branches setup of our \name network versus other architectural setups under unheard sound settings} 
    \resizebox{0.8\linewidth}{!}{
    \begin{tabular}{l|ccccccc}
    \toprule
    \textbf{Architecture} & \textbf{Success $\uparrow$} & \textbf{SPL $\uparrow$} & \textbf{SNA $\uparrow$} & \textbf{DTG $\downarrow$} & \textbf{SWS $\uparrow$} \\    \hline
    \name + Language-based Policy Module  (2-branch) & 31.1 & 20.6 & 16.4 & 8.6 & 21.3 \\
    \name + trajectory forecasting (2-branch) & 27.7 & 19.9 & 17.4 & 9.1 & 15.8 \\
    \name + question branch w.o. oracle instructions (2-branch) & 27.5 & 19.4 & 16.9 & 9.4 & 16.5 \\
    \name + question branch w. oracle instructions (2-branch) & 36.7 & 26.6 & 23.1 & 8.0 & 24.7 \\
    \name (ours) (3-branch) & \textbf{42.0} & \textbf{30.0} & \textbf{26.5} & \textbf{7.6} & \textbf{30.9}  \\ \hline
    \name (ours) (3-branch) + GT actions & \textbf{49.7} & \textbf{37.3} & \textbf{32.7} & \textbf{6.7} & \textbf{37.2} \\
    \bottomrule
    \end{tabular}
    }
    \label{tab:branch_ablation}
\end{table*}

\begin{figure}[ht]
\begin{center}
  \includegraphics[width = \linewidth]{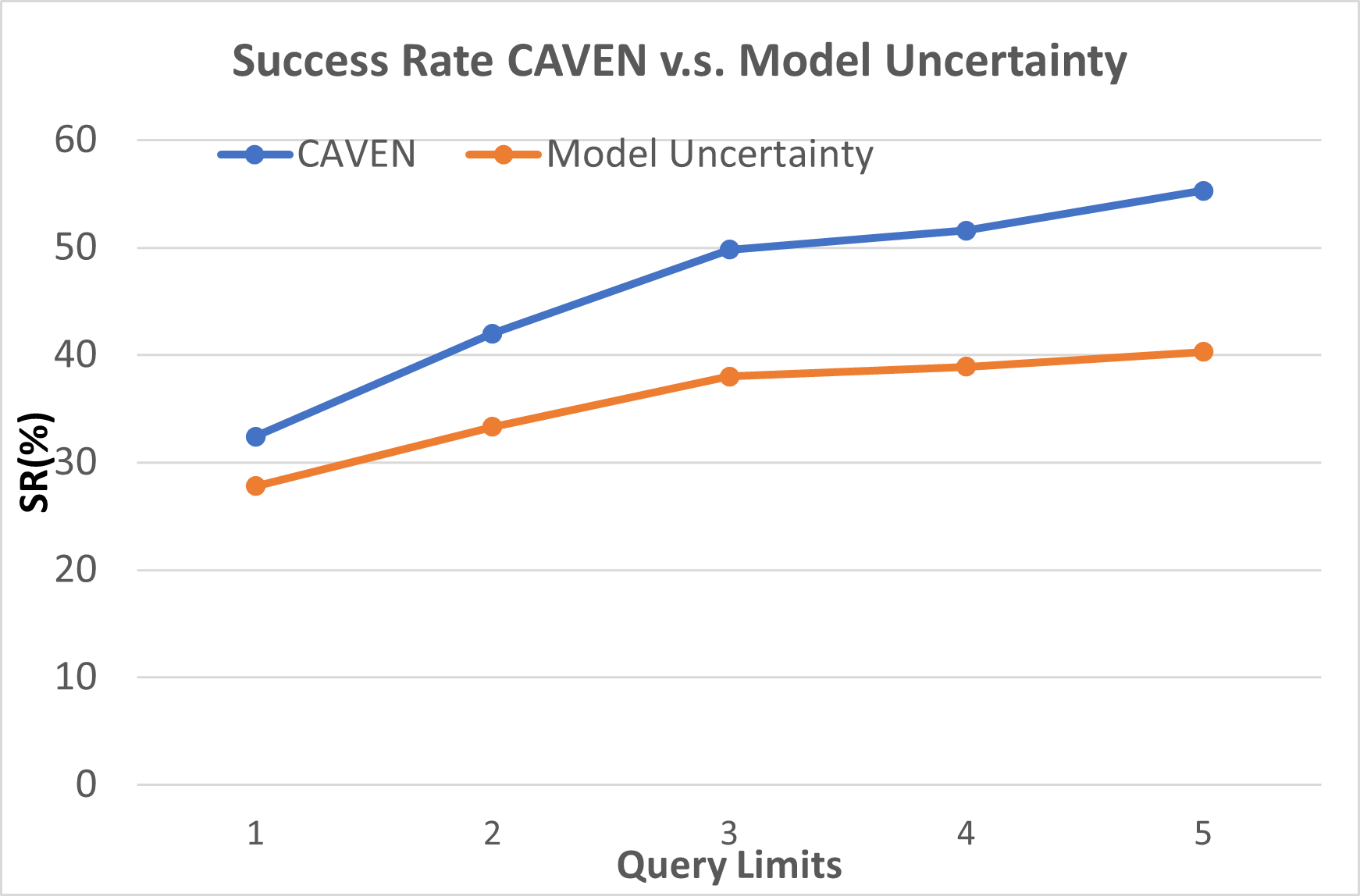} 
  \caption{Plot showing performance, as measured by Success Rate (SR), against varying number of queries for \name against Model Uncertainty.}
  \label{fig:query_limit_SR}
  \end{center}
\vspace{-7mm}
\end{figure}

\begin{figure}[ht]
\begin{center}
  \includegraphics[width = \linewidth]{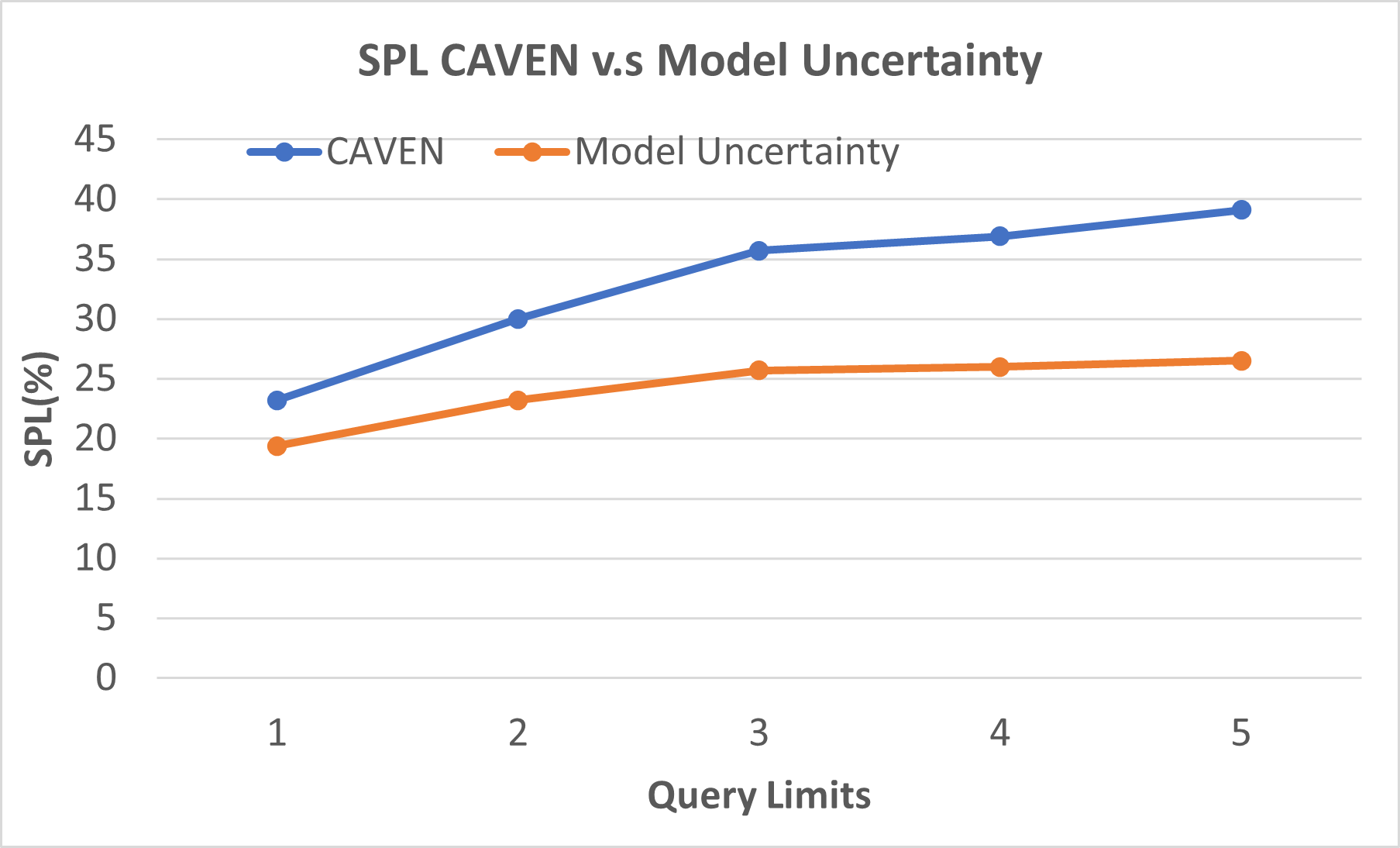} 
  \caption{Plot showing performance, as measured by SPL, against varying number of queries for \name against Model Uncertainty.}
  \label{fig:query_limit_SPL}
  \end{center}
\vspace{-7mm}
\end{figure}

\begin{figure}[ht]
\begin{center}
  \includegraphics[width = \linewidth]{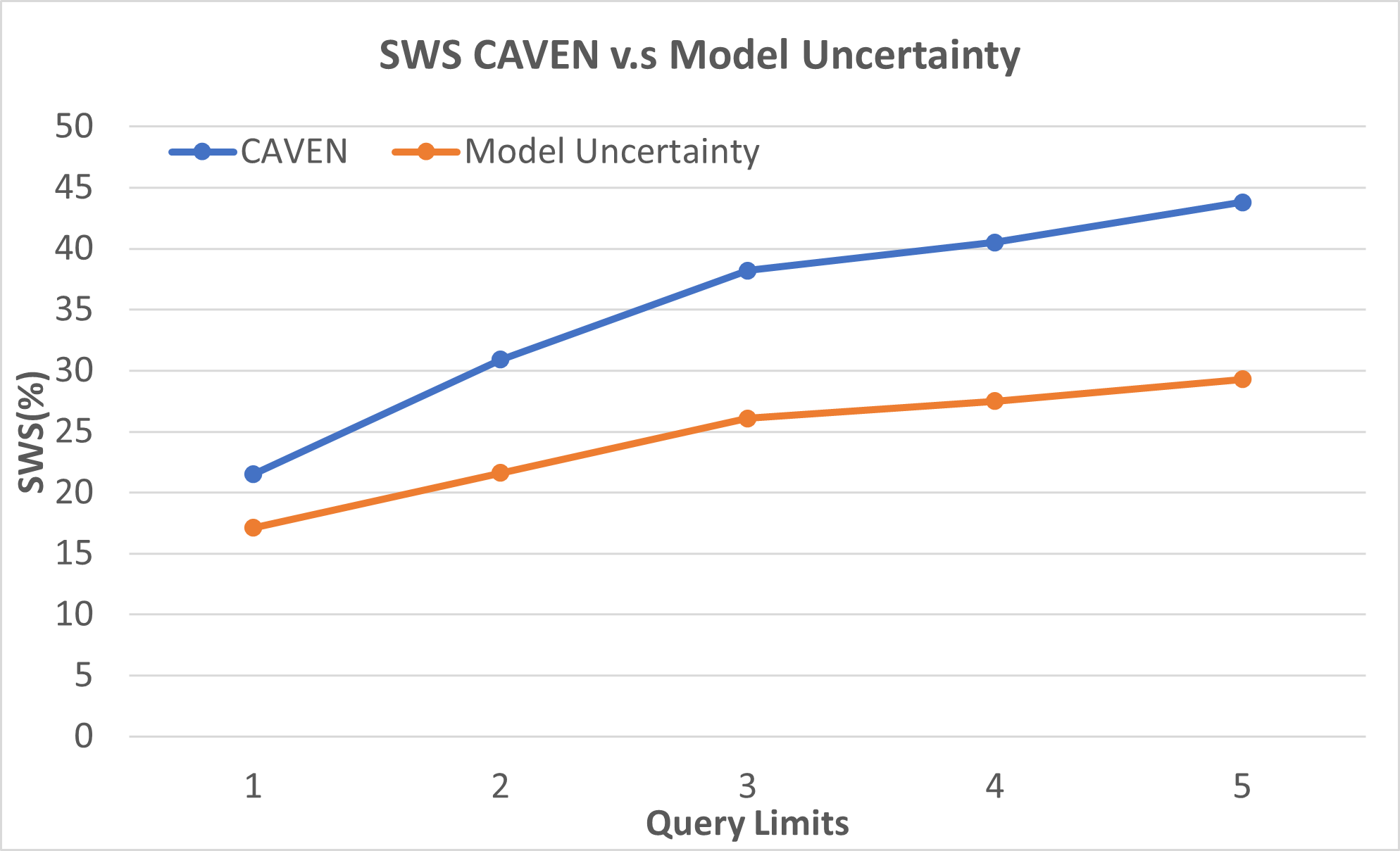} 
  \caption{Plot showing performance, as measured by SWS, against varying number of queries of \name against Model Uncertainty.}
  \label{fig:query_limit_SWS}
  \end{center}
\vspace{-7mm}
\end{figure}

\noindent \textbf{Performance of varying query limit:}
We also study how the navigation performance is impacted when varying the query limits. We evaluate our proposed \name  with the model uncertainty baseline under unheard sound setting by varying query limits from 1 to 5, and plot the Success Rate (SR), SPL and SWS metrics. As shown in Fig. \ref{fig:query_limit_SR}, \ref{fig:query_limit_SPL}, \ref{fig:query_limit_SWS}, the performance of both \name and the model uncertainty method increases as the query limits increase, however, as the query limits become larger, the performance gradually saturates. It is also noteworthy that the gap between the performance of \name and model uncertainty enlarges as the query limits increase underscoring the advantages of conversing with the oracle (or human) while the model uncertainty technique selects a more conservative strategy to select whether or not to query.

\subsection{A5: Implementation Details}
\label{sec:arch}
In this section, we lay out the implementation details of \name's architecture, followed by the details of its training scheme.

\subsection{Architectural Details}

We use a 3-branch architecture and a memory module for implementing the Selector Policy $\pis$.  We set the agent's budget for seeking direct navigation instructions (i.e., total number of times allowed for wrong questions or language-based policy) to 2 per episode.
When queried directly, the oracle's language-policy module takes the oracle actions for the next 4 steps to compute the goal vector. The visual features are obtained by extracting 12 views from the agent's location spanning full 360 degrees, with each view separated by 30 degrees, and are derived from the RGB images of size $640 \times 480$. The ego-centric occupancy maps of size $31 \times 31$ are computed by transforming the depth images from camera coordinates to world coordinates using the camera parameters provided by the simulator. For the question policy branch, $\piquestion$, the \emph{QuestionNet} and the \emph{FollowerNet} share the same architecture as the language-based policy module, while the trajectory forecasting module predicts the next 4 steps using the occupancy encodings of the current surroundings of the agent and a goal vector. 

The \textbf{Bi-directional Question-Answer Policy Module} is one of the centerpieces in the design of our \name framework that enables the agent to pose questions to the oracle, if it is uncertain about its next steps and receive responses in natural language. Below, we elaborate on the architectural details of each of its components.

\begin{figure}[t]
\begin{center}
  \includegraphics[width = \linewidth]{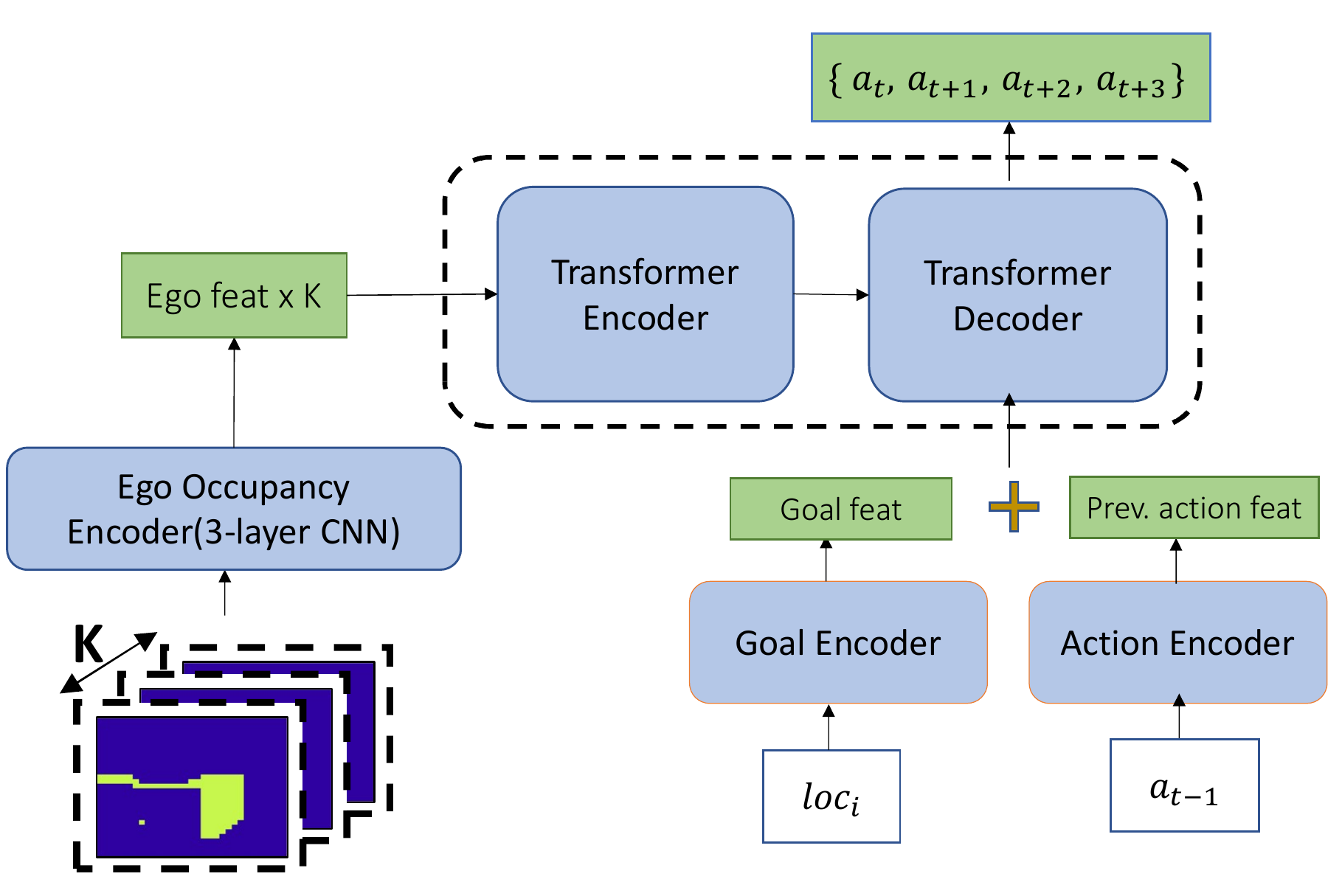} 
  \caption{Architecture of~\emph{TrajectoryNet}.}
  \label{fig:trajectory_net}
  \end{center}
\vspace{-7mm}
\end{figure}

\begin{figure*}[t]
\begin{center}
  \includegraphics[width=0.85\linewidth, trim={0cm 0cm 0cm 0cm}, clip]{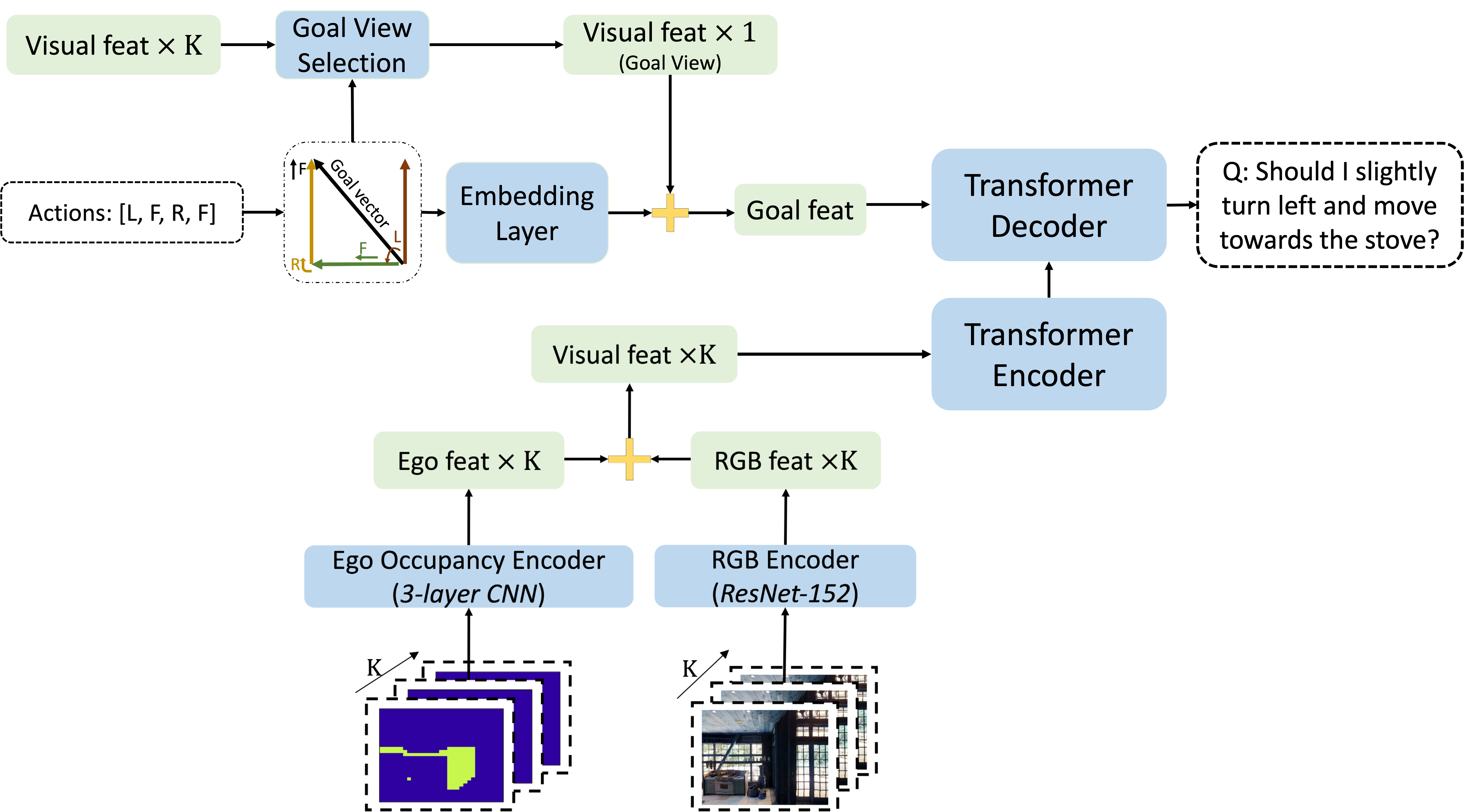} 
  \caption{Architecture of \emph{QuestionNet}.}
  \label{fig:question_net}
  \end{center}
\end{figure*}

\noindent \textbf{TrajectoryNet:} As shown in Fig.\ref{fig:trajectory_net}, the \emph{TrajectoryNet} is a transformer~\cite{vaswani2017attention}-based encoder-decoder network that forecasts the next four steps the agent could take. Its encoder takes as inputs the feature vectors corresponding to the four egocentric occupancy map images, corresponding to the four views spanning the $360^{\circ}$ surroundings of the \name agent (in counter-clockwise order), each separated by $90^{\circ}$. The feature vectors are extracted using a 3-layer Convolutional Neural Network (CNN) which takes as input egocentric occupancy maps of resolution $31 \times 31$ and outputs a flattened 64D vector. The output of the CNN is then projected to 128D using a Multilayer Perceptron (MLP), before being added to a standard sinusoidal position encoding, as is common practice~\cite{vaswani2017attention}. 

On the decoder side, the inputs are 64D vectors derived from a concatenation of a 32D encoding of the audio-goal vector and a 32D encoding of the action taken in the previous step while the output is a prediction of an action step. The space of possible actions is four (viz., move\_forward: 0, turn\_left: 1, turn\_right: 2, stop: 3). The decoder predicts the set of four successive steps, autoregressively, for every invocation of the bi-directional language module both during training and test. 

Both the transformer encoder and decoder are 1-layer multi-head attention networks with 4 heads and a hidden-state size of 128D. The \emph{TrajectoryNet} is trained via the cross-entropy loss with teacher forcing. During inference, we use greedy decoding (i.e., argmax over the predicted action probabilities) to get the action for the next step. 

\noindent \textbf{QuestionNet:} As shown in Fig. ~\ref{fig:question_net}, the \emph{QuestionNet} is also a transformer~\cite{vaswani2017attention}-based encoder-decoder network and is tasked with synthesizing the question to be posed to the oracle (or human). The encoder takes as input the visual features corresponding to the $K=12$ views of the entire surroundings of the agent (in a clockwise order, with each view separated by $30^{\circ}$). The visual features corresponding to each of these views are 512D, obtained by projecting the concatenated 2048D RGB image features encoded via a ResNet-152 network pretrained on the ImageNet dataset~\cite{he2016deep} and the 64D egocentric occupancy map features encoded through the pretrained 3-layer CNN encoder from the \emph{TrajectoryNet} module.

The decoder takes as input the `Goal feat' along with the output hidden states of the encoder. To obtain `Goal feat', we first compute `Goal vector', a 3D vector pointing from the agent's current location to the goal location, from a sequence of actions. As illustrated in Fig. ~\ref{fig:question_net}, an action sequence $[L, F, R, F]$ represents the actions of `Turn Left', `Move Forward', `Turn Right' and `Move Forward' in order. In the simulator, `Move Forward' explicitly means going forward by 1m, while `Turn' means turning by 90 degrees.  Then, we select the view which is closest to the direction of the `Goal vector' from $K=12$ views, and retrieve the visual features of the view as `Visual feat (Goal View)', a single 512D vector. On the other branch, the goal vector is fed through an embedding layer to obtain a feature vector of 96D, whose output is concatenated with `Visual feat (Goal View)' and then are projected through a linear layer to obtain the final `Goal feat'.

The decoder autoregressively decodes the instruction tokens. For word embeddings, we either use 300D Glove embeddings~\cite{pennington2014glove}, pretrained on the Wikipedia dataset and then projected using a single layer MLP to 512D or directly extract 512D pretrained CLIP embeddings~\cite{radford2021learning}. The word embeddings are kept fixed throughout training. The vocabulary size of the instruction dataset is limited to $1625$ after filtering out the uncommon words. Both the encoder and decoder are multi-head attention networks with a 512D hidden state with the number of heads set to 1. Number of layers in the encoder and decoder are set to 1 and 3 respectively. The model is trained with the cross entropy loss on word tokens in a teacher forcing manner. During inference, greedy decoding (using argmax over the space of output tokens) is applied to generate the questions.

\begin{figure}[t]
\begin{center}
  \includegraphics[width = \linewidth]{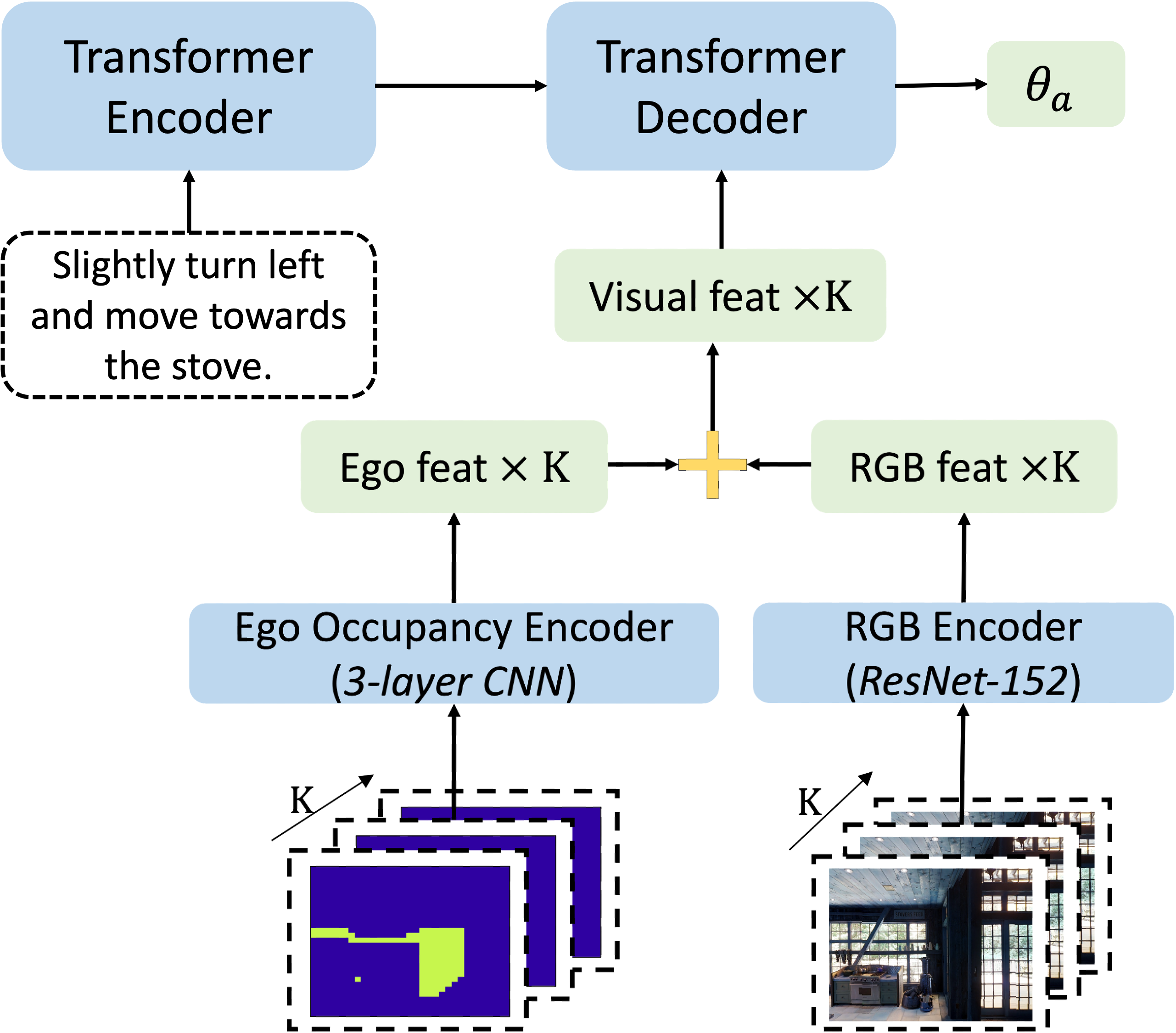} 
  \caption{Architecture of \emph{FollowerNet}}
  \label{fig:follower_net}
  \end{center}
\vspace{-7mm}
\end{figure}

\noindent \textbf{FollowerNet:} Fig.~\ref{fig:follower_net} shows the architectural details of the \emph{FollowerNet}. This module too follows a transformer-based~\cite{vaswani2017attention} encoder-decoder architecture where the encoder takes as input the embeddings of the generated question/instruction, while the decoder takes the visual features corresponding to the four views (each separated by 90 degrees and in a clockwise order) as input, besides the output of the encoder. The visual feature encoder of this network is shared with the \emph{QuestionNet} module. 

Both the encoder and the decoder are 2-layer multi-head attention network with the number of heads being 4 and hidden-size being 512D. The output will be decoded to a discrete direction angle $\theta_a$ (one of 12 equally dividing sectors of a full circle, each spanning 30 degrees) by flattening the last layer of decoder and passing it through a single layer MLP. If the decoded direction is within $30$ degrees error of oracle's estimated direction, then response is 'yes', otherwise 'no'.

The \emph{FollowerNet} is trained using the cross entropy loss on its output. During inference, this module outputs a token which is mapped back to 1 out of the 12 views on the circle.

\subsection{Training Details} 
Since the parameter-intensive transformer architecture~\cite{vaswani2017attention} is at the core of several of our modules, we warm start their training before incorporating them into our Partially Observable Semi-Markov Decision Process (POSMDP) framework, so as to avoid the risk of over-fitting.

\noindent \textbf{Bi-directional Question-Answer Policy Module:} 
The \emph{Bi-directional Question-Answer Policy Module} consists of: (i) the \emph{TrajectoryNet}, (ii) the \emph{QuestionNet}, and (iii) the \emph{FollowerNet}. We pre-train each of these modules to provide better initialization to these networks, before training them jointly in the POSMDP setup. The \emph{TrajectoryNet} is pre-trained using the sub-trajectories and goals supplied with the episodes in the SoundSpaces~\cite{chen2020soundspaces} dataset, following the same train/val/test splits as Chen \etal~\cite{chen2020soundspaces}. We use the sub-instructions provided by the Landmark-RxR dataset~\cite{NEURIPS2021_0602940f}, which contains 150k well-annotated sub-trajectories and their corresponding language sub-instructions grounded on scenes captured using the Matterport3D simulator, to pre-train the \emph{QuestionNet} and \emph{FollowerNet} modules so that there is synergy between the two before integrating them into a single network and training them in a joint fashion. After the pre-training stage, to make the FollowerNet adapt to the grid strcuture of SoundSpaces simulation environment, we use our synthesized AVN-Instruct dataset (described in AVN-Instruct Dataset section~\ref{sec:avn}) to train the \emph{QuestionNet} and \emph{FollowerNet} in a joint and self-supervised manner. The \emph{QuestionNet} is trained using cross entropy loss with teacher forcing on ground truth instructions, while the \emph{FollowerNet} is optimized using cross entropy loss with respect to the ground truth direction angle. During the pre-training stage using Landmark-RxR dataset, we use the Adam Optimizer with default settings and set the learning rate to be 1e-4 for both QuestionNet and FollowerNet. After that, in stage 2 joint training on AVN-Instruct dataset, we set the learning rate of QuestionNet be 1e-6 to avoid divergence from pre-trained model, while keep the learning rate for FollowerNet as 1e-4 to achieve better adaptation to the SoundSpaces environment.

\noindent \textbf{Policy Modules:} \name is trained using our proposed budget-aware reinforcement learning (RL) policy framework under the POSMDP settings. The \emph{Selector Policy} $\pi_s$ is trained from scratch within the RL setup. However, each of the three option policies is pre-trained before being integrated into the RL setup. (i) The learning of the \emph{audio-visual navigation policy}, $\pigoal$ entails an off-policy and an on-policy training regime. For the off-policy training, we follow the SAVi setup~\cite{chen2021semantic}. We then follow a two-stage training process for the on-policy training akin to Chen \etal~\cite{chen2020soundspaces}, the first of which does not use the memory, while the second one does. (ii) The language-based policy, $\pivln$, is initialized with the pre-trained \emph{FollowerNet} module, discussed above. Post this pre-training, it is integrated into the RL setup for fine-tuning. (iii) The \emph{bi-directional question-answer} policy, $\piquestion$ can be learned by either freezing or fine-tuning the different components of the corresponding policy module branch (consisting of the \emph{TrajectoryNet}, and \emph{QuestionNet} at the agent's side and the \emph{FollowerNet} at the oracle's side) using the actions sampled during navigation. For the trajectory forecasting module, we evaluate the accuracy of accumulated actions over four steps against the shortest path to the audio-goal.

\begin{figure}[t]
  \centering
  \includegraphics[width=\linewidth]{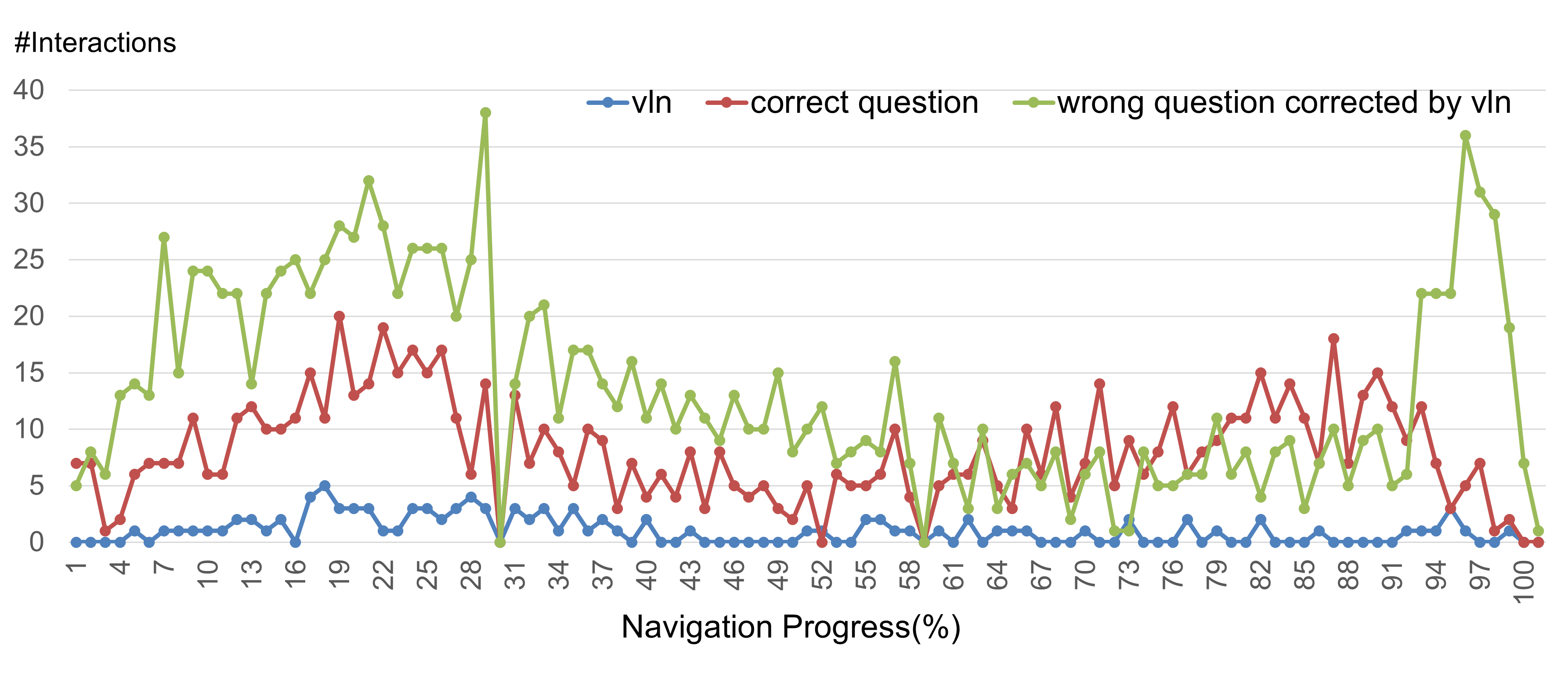}
  \caption{Distribution of queries versus the timesteps averaged over episodes (`vln' denotes language-based policy)}
  \label{fig:qdist}
\end{figure}

\noindent \textbf{Analysis of When Interactions Happen:} To understand when the agent is likely to interact, we keep track of the interaction status in each episode in the test set and record whether the questions being asked are correct or not. We observe from Figure~\ref{fig:qdist} that the interaction distribution shows bimodal patterns. Specifically, the agent tends to have a high number of interactions towards the first one-third of the episodes, and another peak is observed towards the end. This is intuitive since when the agent begins to navigate, it is far away from the audio-goal. Moreover, the sound is intermittent and can disappear at any time. So it is essential that the agent seeks help from the oracle. Again, when the agent is close to the audio-goal, it needs to invoke the `stop' action near the goal. Invoking `stop' action correctly is difficult to learn for an RL agent as the agent needs to be certain where to invoke the `stop' action (else it may get a penalty). Hence, the model may learn to seek direct instructions from the oracle in such cases. Further, note that the agent poses more correct questions (see Figure~\ref{fig:qdist}) than incorrect ones and also asks fewer questions when it is certain about navigation steps.

\subsection{A6: Details of our Newly Introduced Metrics}
\label{sec:metrics}
We introduce two new metrics for assessing the accuracy of navigation, which factor in the number of interactions with the oracle. These are: (i) $SNI$ (Success rate weighted by the inverse Number of language Interactions), and (ii) $SNO$ (Success rate weighted by the inverse Number of Oracle instructions). These are defined in more details as follows:
\subsection{SNI}
SNI assesses the accuracy of a model's navigation per language interaction with the oracle and is mathematically defined as:
\[
\text{SNI} = \frac{\text{SR}}{(N_{ques} + N_{l}) / N}
\]
where SR is the success rate, $N$ is total number of episodes in test set, $N_{ques}$ is the total number of times the bi-directional language module is invoked, while $N_{l}$ denotes the total number of times the agent directly seeks for language-based instruction. This metric thus penalizes a model if it interacts with the oracle too frequently.

\subsection{SNO}
SNO evaluates the navigation performance of a model per direct instruction sought from the oracle and is mathematically defined as:
\[
\text{SNO} = \frac{\text{SR}}{(N_{l} + N_{ques\_wrong}) / N}
\]
where SR is the success rate, $N$ is total number of episodes in test set, $N_{l}$ is the total number of times the agent queries the oracle for navigation instruction directly, while $N{ques\_wrong}$ represents the total number of times the oracle says 'No' to a question posed by the agent and consequently provides instructions for the agent to take. $SNO$ is thus a slightly more relaxed version of the $SNI$ score and penalizes a model only if it receives direct language-based navigation instruction from the oracle.

\begin{figure}[t]
\centering
  \includegraphics[width=\linewidth]{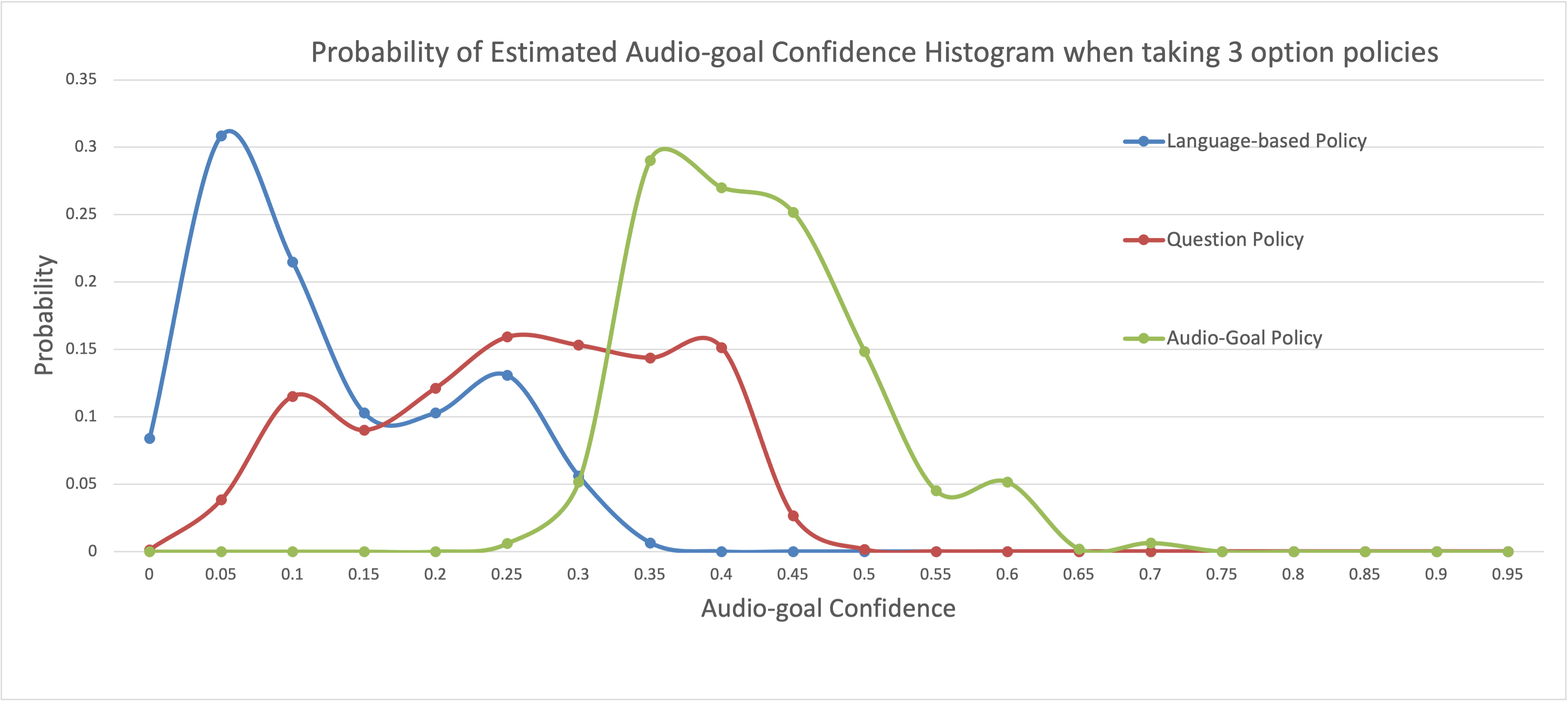}
  \caption{Distribution of estimated audio goal confidence when each policy is invoked.}
  \label{fig:gdist_supp}

\end{figure}

\subsection{A7: Analysis of Policy Dynamics} 

In an ideal scenario, a navigating agent should navigate autonomously when it is certain about its navigation steps and ask questions when it has some clue regarding the potential direction. Again, the agent should rely on only oracle instructions when it is completely uncertain about the next steps to take. This can be achieved by modeling the uncertainty of the agent and setting up an operating threshold in an ad-hoc manner. However, modeling the uncertainty of the agent is computationally expensive and manually defined operating range is likely to be sub-optimal. Instead, in our empirical study, we observe that our POSMDP setup allows us to learn policies that implicitly captures the desired characteristics of uncertainty-based navigation without the requirement of directly modeling uncertainty. To study the situations when the agent invokes the various navigation policies, we record the confidence of audio-goal estimated by the selector policy $\pi_s$, when each of the option policies is invoked and compute its distribution using all test set episodes. As shown in Figure~\ref{fig:gdist_supp}, the audio-goal is invoked when the agent is highly confident. The language-based policy is invoked when agent's confidence is low. It is noteworthy that the question policy is invoked more often when the agent is moderately confident. Though it potentially risks being penalized by asking wrong questions, it benefits from seeking confirmation from the oracle using its own audio-visual cues to help alleviate the navigation uncertainty, thus facilitating efficient navigation.

\subsection{A8: Limitations}
There are two key limitations of our \name framework: (i) Firstly, the interactions between the agent and the oracle all happen in English which prevents its applicability to a more wider range of non-English speaking users at the moment, (ii) secondly, the \emph{QuestionNet} and the \emph{FollowerNet} are both pre-trained on the Landmark-RxR dataset~\cite{NEURIPS2021_0602940f} -- grounded on only a portion of observations (sparser than SoundSpaces) in MatterPort3D simulator -- which may not cover all categories of objects in the SoundSpaces~\cite{chen2020soundspaces} environment, thereby resulting in potential underfitting of the trained modules.

\bibliography{aaai24}

\end{document}